\documentclass[journal]{IEEEtran}
\usepackage{amsmath,amsfonts}
\usepackage{algorithmic}
\usepackage{algorithm}
\usepackage{array}
\usepackage[caption=false,font=normalsize,labelfont=sf,textfont=sf]{subfig}
\usepackage{textcomp}
\usepackage{stfloats}
\usepackage{url}
\usepackage{verbatim}
\usepackage{graphicx}
\usepackage{cite}
\usepackage[table]{xcolor}
\definecolor{lightblue}{RGB}{230, 240, 255}
\definecolor{blockA}{RGB}{230,240,255}  
\definecolor{blockB}{RGB}{240,255,240}  
\definecolor{blockC}{RGB}{255,245,230}  
\definecolor{cat1}{RGB}{235,245,255}  
\definecolor{cat2}{RGB}{240,255,240}  
\definecolor{cat3}{RGB}{255,245,220}  
\definecolor{cat4}{gray}{0.8}
\usepackage{makecell}
\usepackage{pifont}
\usepackage[table]{xcolor}
\usepackage[colorlinks,linkcolor=blue,citecolor=blue]{hyperref}
\usepackage{hyperref}
\usepackage{multirow}

\begin{document}

\title{Dual-Space Modality Consistency Learning for Universal Cross-Modal Re-Identification}

\author{Yujian Zhao, Yukang Zhao, Hankun Liu, Haoxuan Xu, Bo Li, Hanzi Wang,~\IEEEmembership{Senior Member,~IEEE}, Guanglin Niu
\thanks{Yujian Zhao, Haoxuan Xu, Bo Li and Guanglin Niu are with the School of Artificial Intelligence, Beihang University (E-mail: boli@buaa.edu.cn; beihangngl@buaa.edu.cn). Yuhang Zhao is with the School of Humanities and Social Sciences, Beihang University. Hankun Liu is with the School of Computer Science and Engineering, Beihang University. Hanzi Wang is with the Fujian Key Laboratory of Urban Intelligent Sensing and Computing, School of Informatics, Xiamen University, Xiamen 361005, China (E-mail: hanzi.wang@xmu.edu.cn).

Corresponding author: Guanglin Niu.

This work was partially supported by the National Natural Science Foundation of China (No. U25A20531 and No. 62376016).}
}

\markboth{Journal of \LaTeX\ Class Files,~Vol.~14, No.~8, August~2021}%
{Shell \MakeLowercase{\textit{et al.}}: A Sample Article Using IEEEtran.cls for IEEE Journals}


\maketitle


\begin{abstract}
Cross-modal Re-Identification (ReID) aims to retrieve the same identity across heterogeneous imaging modalities and has been widely studied in visible-infrared person ReID and cross-modal ship ReID. Existing methods have achieved promising performance by learning modality consistency in the spatial embedding space, yet often overlook frequency-domain modality discrepancy, particularly in high-frequency representations that are both highly discriminative and  modality-sensitive. In addition, most approaches are tailored to specific modality settings, limiting their applicability across diverse cross-modal scenarios. To address these challenges, we propose a Dual-Space Modality Consistency Learning (DSMCL) framework for universal cross-modal ReID. Specifically, DSMCL jointly models spatial feature distribution consistency and frequency-domain discriminative consistency. A Spatial Modality Consistency Learning (SMCL) branch performs Gaussian-based feature alignment, while a Frequency-aware Discriminative Consistency Learning (FDCL) strategy regularizes high-frequency representations through identity-aware cross-modal contrastive learning. By jointly capturing modality-specific characteristics and modality-shared identity cues, DSMCL learns robust representations and establishes a unified framework capable of accommodating diverse heterogeneous modality settings. Moreover, DSMCL is a plug-and-play framework that can be readily integrated into existing cross-modal ReID architectures. Extensive experiments on SYSU-MM01, RegDB, LLCM, HOSS-ReID, and CMShipReID across seventeen evaluation protocols show that DSMCL consistently improves multiple representative baselines. Specifically, it improves the Rank1 from 88.2\% to 92.3\% for IDKL on SYSU-MM01 under the Indoor-search Single-shot protocol and from 31.3\% to 38.8\% for TransOSS on HOSS-ReID under the SAR-to-Optical protocol. It also improves the mAP of TransReID from 66.8\% to 73.9\% on CMShipReID under the TIR-to-VIS protocol, demonstrating its effectiveness and generalizability across diverse cross-modal ReID scenarios.



\end{abstract}


\section{Introduction}

\begin{figure}[!t]
\centering
\includegraphics[width=3.5in]{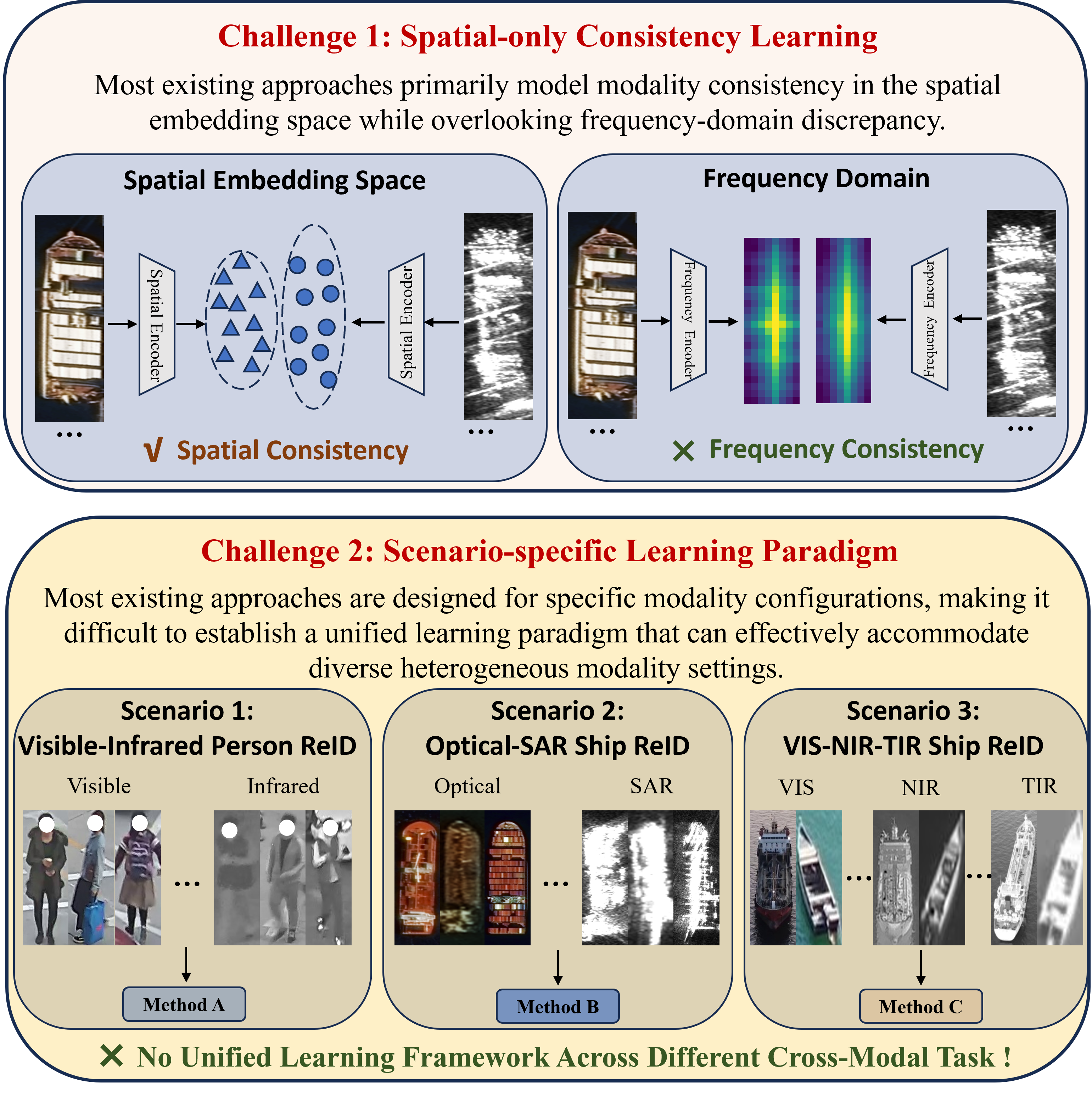}
\caption{Illustration of the limitations of existing cross-modal ReID methods. Existing approaches mainly focus on spatial embedding alignment while often neglecting frequency-domain modality discrepancy. Moreover, most methods are designed for specific modality configurations, hindering the establishment of a unified learning paradigm across diverse heterogeneous modality settings. These challenges motivate our DSMCL framework for addressing the problem of universal and robust cross-modal ReID.
}
\label{fig:intro}
\end{figure}

\IEEEPARstart{C}{ross-modal} Re-Identification (ReID)  aims to match images of the same target captured by heterogeneous sensing modalities \cite{reid-clipreid},\cite{vireid-x},\cite{instructreid},\cite{GI-ReID},\cite{TryHarder},\cite{CCUP},\cite{yuan2026towards}. Owing to its broad applications in intelligent surveillance, maritime monitoring, and multi-sensor security systems, cross-modal ReID has attracted increasing attention in recent years. Compared with conventional single-modality ReID, cross-modal ReID must bridge substantial appearance discrepancies introduced by heterogeneous imaging mechanisms. Typical examples include visible-infrared person ReID \cite{huang2023deep}, \cite{deen}, \cite{yu2026x}, \cite{feng2023shape}, \cite{vireid-MCL}, \cite{vireid-neural}, optical-SAR ship ReID \cite{transoss},\cite{mos},\cite{fan2025smart}, and multi-modal ship ReID involving visible, near-infrared, and thermal infrared imagery \cite{cmshipreid}. Despite the diversity of sensing modalities and application scenarios, all these tasks share a common objective: learning identity-robust representations across heterogeneous modalities. However, the substantial modality discrepancy introduced by different sensing mechanisms often dominates identity-related variations, making robust cross-modal representation learning a fundamental challenge.

Despite the remarkable progress achieved by existing cross-modal ReID methods, two important limitations remain, as illustrated in Fig.~\ref{fig:intro}. First, most existing approaches primarily model modality consistency in the spatial embedding space while overlooking frequency-domain discrepancy. Second, existing methods are typically designed for specific modality configurations, making it difficult to establish a unified learning paradigm that can effectively accommodate diverse heterogeneous modality settings.

To mitigate modality discrepancy, existing methods have explored various strategies, including image translation \cite{yuan2025modality},\cite{dive},\cite{qi2023generative},\cite{pan2024unified},\cite{yuan2026towards}, modality-invariant feature learning \cite{feng2019learning},\cite{icsd},\cite{lu2023learning},\cite{feng2022visible}, distribution alignment \cite{dma},\cite{zhang2025adaptive},\cite{DSFAD},\cite{dsaf}, and contrastive representation learning \cite{li2025propagation},\cite{sun2022not},\cite{zhang2021attend},\cite{csdn}. Despite their promising performance, most existing approaches primarily focus on learning modality consistency in the spatial embedding space \cite{deen},\cite{feng2023shape},\cite{MCJA},\cite{pan2026prototype}. Specifically, they typically model modality discrepancy as feature distribution shift and enforce cross-modal feature alignment through adversarial learning, distribution matching, or feature consistency constraints. Although these methods effectively reduce global distribution discrepancy across modalities, they implicitly assume that modality inconsistency can be sufficiently characterized within the spatial feature space.

\begin{figure}[!t]
\centering
\includegraphics[width=3.5in]{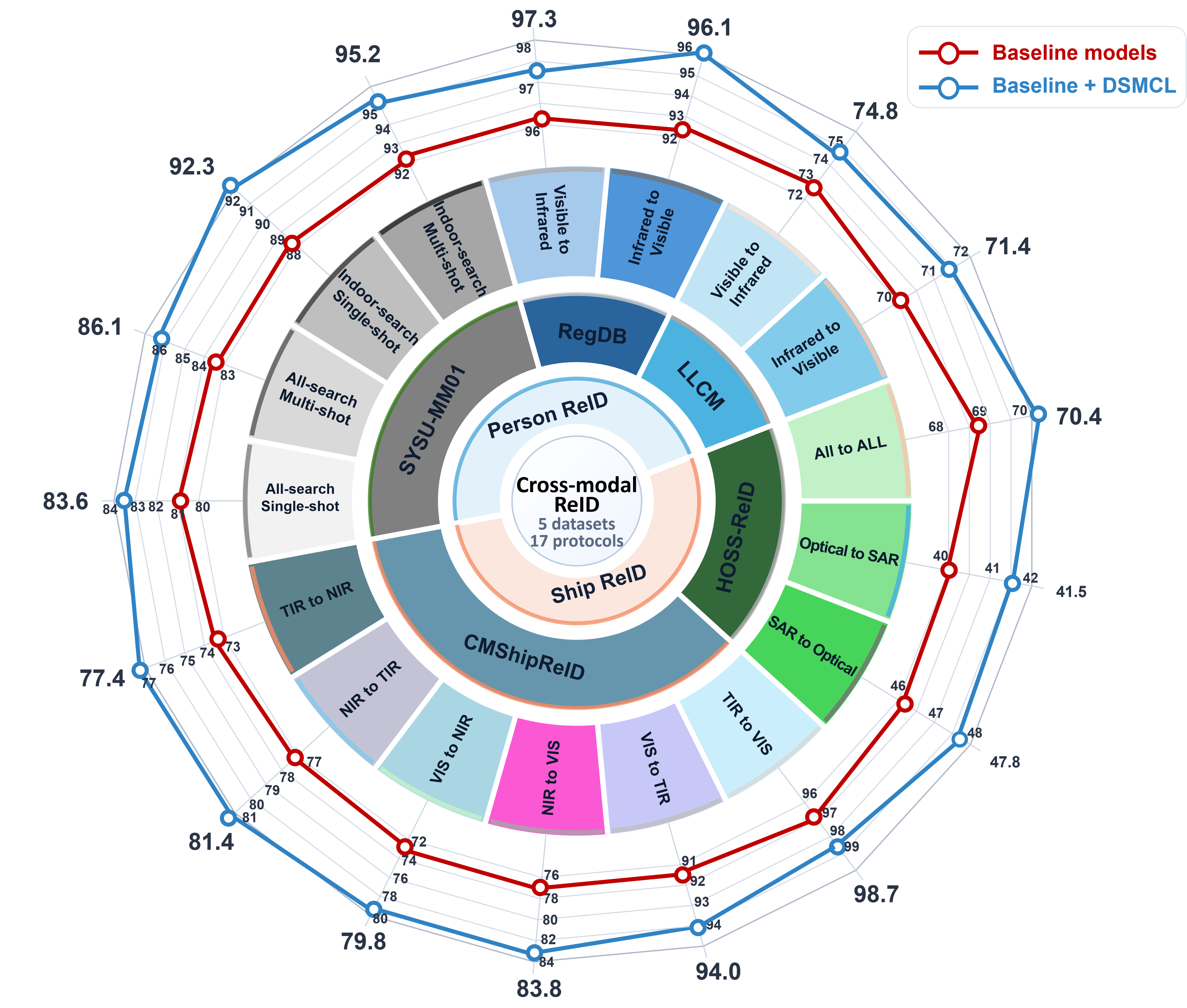}
\caption{Overview of the universal cross-modal ReID evaluation setting and the performance gains brought by DSMCL across five benchmark datasets and seventeen evaluation protocols.}
\label{fig:radar}
\end{figure}

However, heterogeneous imaging mechanisms affect not only spatial feature distributions but also the inherent frequency responses of visual representations \cite{zhou2024general,zheng2025frequency}. For instance, infrared imaging tends to suppress high-frequency texture information due to thermal smoothing effects, whereas SAR imaging often introduces strong high-frequency perturbations caused by speckle noise. Such modality-specific frequency distortions are difficult to characterize solely through spatial-domain alignment. More importantly, modality discrepancy is not uniformly distributed across the frequency spectrum \cite{yang2025freqcross}. While low-frequency components mainly preserve coarse structural semantics that can already be effectively optimized by standard identity supervision, high-frequency components simultaneously contain modality-sensitive distortions and identity-discriminative details \cite{li2024adaptive,MFENet}. Consequently, directly applying naive spatial-domain consistency learning may fail to fully and effectively model the most challenging cross-modal discrepancies.

A straightforward solution is to perform frequency-domain alignment across heterogeneous modalities. However, indiscriminately minimizing spectral discrepancies may suppress identity-discriminative patterns alongside modality-specific distortions, leading to over-smoothed representations and degraded retrieval. Therefore, an important question naturally arises: how can we effectively reduce high-frequency modality discrepancies while preserving discriminative identity cues.


To address this issue, we propose a Dual-Space Modality Consistency Learning (DSMCL) framework for universal cross-modal ReID. Different from existing methods that model modality consistency solely in the spatial embedding space, DSMCL jointly explores modality consistency from both the spatial feature space and the frequency representation space. Specifically, we first employ Spatial Modality Consistency Learning (SMCL) to reduce class-wise feature distribution discrepancy across heterogeneous modalities through Gaussian-based distribution alignment. Furthermore, we introduce Frequency-aware Discriminative Consistency Learning (FDCL), which explicitly models high-frequency representations through identity-aware cross-modal contrastive consistency learning. By jointly optimizing spatial consistency and frequency-aware discriminative consistency, DSMCL learns identity-consistent yet modality-robust representations for cross-modal retrieval.

The second challenge concerns the universality of existing cross-modal ReID paradigms. Most current approaches are developed and validated under a specific modality configuration, while their applicability to substantially different cross-modal scenarios remains largely unexplored.

For example, visible-infrared ReID methods are mainly designed for RGB-IR benchmarks such as SYSU-MM01 \cite{sysumm01}, RegDB \cite{regdb}, and LLCM \cite{deen}, while cross-modal ship ReID methods typically focus on optical-SAR \cite{transoss} or visible-NIR-TIR maritime scenarios \cite{cmshipreid}. Owing to the substantial differences among sensing mechanisms, these methods are often built upon modality-specific observations, assumptions, or optimization strategies tailored to a particular cross-modal setting. Although such designs have achieved remarkable performance on their target benchmarks, whether the underlying modeling paradigms can consistently benefit substantially different cross-modal scenarios remains largely unexplored.

In practical applications, however, cross-modal retrieval systems are increasingly deployed in heterogeneous sensing environments involving diverse sensor combinations. For instance, intelligent surveillance systems commonly integrate visible and infrared cameras to enable all-weather pedestrian retrieval, whereas maritime monitoring platforms often simultaneously employ optical, SAR, near-infrared, and thermal infrared sensors to improve target perception under complex environmental conditions. Despite the diversity of sensing modalities and application scenarios, these tasks share a common objective: learning identity-consistent representations that remain robust to modality variations. Therefore, developing a unified representation learning framework that can effectively accommodate different cross-modal settings has become an important yet underexplored research direction.

To this end, the proposed DSMCL is designed as a universal cross-modal representation learning framework. Instead of relying on modality-specific assumptions, DSMCL explicitly models modality consistency from both the spatial feature space and the frequency representation space, thereby providing a more general learning paradigm for heterogeneous sensing modalities. As illustrated in Fig.~\ref{fig:radar}, aiming to comprehensively evaluate its effectiveness and applicability, we conduct extensive experiments on multiple cross-modal person ReID and ship ReID benchmarks, including SYSU-MM01 \cite{sysumm01}, RegDB \cite{regdb}, LLCM \cite{deen}, HOSS-ReID \cite{transoss}, and CMShipReID \cite{cmshipreid}. Experimental results demonstrate that DSMCL consistently achieves superior performance across substantially different modality combinations, validating the effectiveness and superiority of dual-space modality consistency learning for robust universal cross-modal ReID.


This paper substantially extends our previous conference work MOS \cite{mos} in three respects. First, we replace the diffusion-based cross-modal data generation and feature fusion modules in MOS with the training-only FDCL strategy to eliminate their substantial inference overhead and support practical deployment. Second, we extend the spatial-domain consistency learning in MOS to a dual-space formulation by introducing FDCL, which adaptively extracts high-frequency representations and enforces identity-aware cross-modal contrastive regularization. Third, we generalize the framework from Optical-SAR ship ReID to universal cross-modal ReID scenarios, including visible-infrared person ReID and multi-modal ship ReID, demonstrating stronger robustness and generalization capability across heterogeneous modalities. 


In summary, our main contributions are as follows: 
\begin{itemize}

    \item We develop a Dual-Space Modality Consistency Learning (DSMCL) framework that jointly regularizes class-wise spatial feature distributions and frequency-domain discriminative representations for cross-modal ReID. And DSMCL can be plugged into existing methods during training without introducing additional inference cost.

    \item We introduce a Spatial Modality Consistency Learning (SMCL) method to align class-wise spatial feature distributions across modalities. By modeling identity-specific modality distributions with Gaussians and minimizing their Wasserstein-2 distance, SMCL aligns both the mean and variance of spatial representations.

    \item We propose a Frequency-aware Discriminative Consistency Learning (FDCL) strategy that adaptively extracts high-frequency representations and enforces identity-aware cross-modal contrastive regularization, complementing spatial distribution alignment with modality-invariant discriminative details.

    \item Extensive experiments on five cross-modal person and ship ReID benchmarks  demonstrate that DSMCL consistently improves multiple baselines and achieves competitive or superior performance. Ablation studies further validate the effectiveness of each component.

\end{itemize}

\section{Related Work}

\subsection{Cross-modal ReID}
Cross-modal Re-Identification (ReID) aims to match images of the same target captured by heterogeneous sensing modalities. Compared with conventional single-modality ReID, cross-modal ReID faces additional challenges caused by substantial appearance discrepancy introduced by different sensing mechanisms \cite{CCUP,TryHarder, vireid-part, versReID}. Among existing research directions, visible-infrared person ReID (VI-ReID) is one of the most extensively studied topics \cite{survey_ye}. Existing methods can generally be divided into two categories.

The first category focuses on learning modality-invariant representations. These methods attempt to reduce modality discrepancy directly in the feature space by learning identity-discriminative yet modality-robust representations. For example, BIT \cite{bit} proposes a method named Bi-directional Interaction Transformation that adopts a matching-based strategy that explicitly models the interaction between visible and infrared image pairs. DEEN \cite{deen} proposes a diverse embedding expansion network that effectively generates diverse embeddings to learn the informative feature representations and reduce the modality discrepancy. HOH-Net \cite{HOH-Net} introduces a high-order structure learning method to explore the high-order relationships of the feature nodes and exploit a hierarchical middle-feature agent learning loss to reduce the modality discrepancy. ICSD \cite{icsd} introduces identity-compensated style distillation network that enforces cross-modality style consistency and enhances the discriminative power of modality-invariant features. CMSD \cite{cmsd} proposes cross-layer semantic distillation and intra-layer alignment distillation method to guide low-level features in learning more reliable modality invariant representations. These methods have significantly improved visible-infrared retrieval performance by learning discriminative modality-invariant feature representations. 

The second category focuses on image translation and modality compensation. Instead of directly reducing the inherent modality discrepancy in the feature space, these methods attempt to narrow the appearance gap at the image level by generating modality-compensated samples. MTRL \cite{yuan2025modality} proposes a novel framework via modality-transition representation learning  with a middle generated image as a transmitter from visible to infrared modals. DiVE \cite{dive} proposes a data generation framework that automatically obtains massive RGB-IR paired images while preserving identity information by decoupling identity and modality, thereby improving the performance of existing models. Pan \emph{et al}.~\cite{pan2024unified} adapts the conditional diffusion model for multi-modal image generation and employs the adversarial training to measure the distance from the visible or infrared modality to the middle modality. AGPI$^2$ \cite{AGPI} introduces an adaptive generation of privileged intermediate information training approach to adapt and generate a virtual domain that bridges discriminative information between the visible and infrared modalities. By translating visible images into infrared style or vice versa, these approaches can effectively alleviate potential appearance discrepancy caused by heterogeneous sensing mechanisms. However, they usually require additional generative models and introduce significant computational overhead.

Beyond pedestrian retrieval, cross-modal ReID has recently been extended to maritime scenarios. Wang \emph{et al}.~\cite{transoss} establish the first optical-SAR ship ReID benchmark, termed HOSS-ReID, and propose a dedicated cross-modal ship ReID framework named TransOSS that incorporates a dual-head tokenizer and a modality-shared transformer encoder to bridge the substantial modality discrepancy between optical and SAR imagery. MOS \cite{mos} further extends TransOSS by introducing a modality consistency learning framework that mitigates the optical-SAR modality gap through robust representation learning during training and cross-modal feature fusion during inference. Xu \emph{et al}.~\cite{cmshipreid} introduces CMShipReID, a large-scale cross-modal ship ReID benchmark comprising visible, near-infrared, and thermal infrared images collected by an unmanned aerial vehicle, providing a challenging platform for multi-modal maritime retrieval. Fan \emph{et al}.~\cite{fan2025smart} introduce SMART-Ship, a multi-modal ship ReID benchmark comprising optical, SAR, and panchromatic (PAN) imagery, which provides a challenging benchmark for heterogeneous maritime target retrieval. More recently, several preliminary studies have further explored cross-modal ship retrieval under more complex sensing conditions \cite{xian2025beyond, chen2026sdf}. These studies demonstrate the growing interest in maritime cross-modal retrieval and highlight the necessity of developing more robust and universally applicable representation learning frameworks across diverse modality configurations.

Despite these advances, most existing cross-modal ReID methods are still developed and evaluated under specific modality settings. Person ReID methods are typically designed for RGB-IR scenarios, whereas ship ReID methods mainly focus on optical-SAR or visible-NIR-TIR settings. As a result, the generalization capability of current approaches across substantially different sensing modalities remains insufficiently explored. In contrast, this work aims to develop a robust representation learning framework that can effectively accommodate diverse heterogeneous modality settings and establish a unified learning paradigm for universal cross-modal ReID.
\subsection{Modality Alignment Learning}

Modality alignment learning aims to reduce the discrepancy between heterogeneous sensing modalities by encouraging modality-consistent representations. As one of the most widely adopted paradigms in cross-modal ReID, modality alignment has been extensively studied in both visible-infrared person retrieval and maritime cross-modal retrieval scenarios. Existing approaches generally perform modality alignment either at the image level or at the feature representation level.

Image-level alignment methods attempt to narrow the modality gap by translating samples from one modality into another through generative models \cite{yuan2025modality}. Early studies mainly relied on generative adversarial networks to synthesize modality-compensated samples, while more recent approaches employ diffusion-based generation frameworks to construct richer cross-modal supervision signals \cite{dive},\cite{pan2024unified}. Qi \emph{et al}.~\cite{qi2023generative}  proposes a generative-based cross-modality image fusion strategy to  generate high-quality cross-modality paired images and fuse the information of the two modalities. By reducing appearance discrepancy at the image level, these methods provide auxiliary training samples that facilitate subsequent representation learning. Nevertheless, image translation approaches usually require additional generative networks and may suffer from semantic drift or inconsistency introduced during the generation process.

Feature-level alignment methods directly reduce modality discrepancy in the embedding space without explicitly generating additional samples. AMML \cite{zhang2025adaptive} proposes an adaptive middle-modality alignment learning method that reduces the modality discrepancy via an adaptive middle modality learning strategy at both image level and feature level. MCJA \cite{MCJA} introduces a multi-level cross-modality joint alignment to bridge both the modality and the objective-level gap. DMA \cite{dma} proposes a dual modality-aware alignment model to simultaneously preserve discriminative identity information and suppress the misleading information within a uniform scheme. DSFAD \cite{DSFAD} introduces a novel diverse semantics-guided feature alignment and decoupling network to effectively align identity-relevant features from different modalities. These methods generally enforce cross-modal consistency by aligning feature distributions or enhancing semantic correspondence between heterogeneous modalities. Compared with image-level alignment, feature-level methods are computationally more efficient and have consequently become the dominant paradigm in recent cross-modal ReID research.

Among feature-level alignment strategies, distribution alignment has emerged as an effective solution for reducing modality discrepancy by matching heterogeneous feature distributions \cite{dma,DSFAD}. However, existing methods mainly perform global distribution matching and seldom consider identity-aware class-wise consistency. In contrast, the proposed SMCL explicitly aligns class-wise Gaussian distributions across modalities, thereby achieving more discriminative modality consistency learning.

\subsection{Frequency-domain Representation Learning}

Frequency-domain representation learning has recently attracted increasing attention in computer vision. Compared with conventional spatial-domain representations, frequency-domain representations provide complementary information for characterizing  global image structures, fine-grained textures, and spectral responses \cite{xu2020learning}. Existing studies have demonstrated that frequency information can effectively improve robustness, generalization capability, and representation quality in various vision tasks \cite{qian2020thinking}.

Motivated by these observations, frequency-domain learning has consequently and increasingly  been introduced into person ReID and cross-modal representation learning \cite{MFENet, zhang2023pha}. MFEN \cite{mfen} proposes a multi-frequency expert network that enables multi-frequency modulation and combines different frequencies through a mixture-of-experts method. MFENet \cite{MFENet} introduces a novel multi-frequency embedding network, a feature-level method that operates in the frequency domain through multi-frequency decomposition to learn discriminative and modality-invariant features. FFD-Net \cite{FFD-net} proposes a frequency-driven feature decoupling network that uses frequency domain supervision to further improve cross-modal matching. FDNM \cite{fdnm} introduces a frequency domain nuances mining method to effectively explore the cross-modality frequency domain information. By explicitly modeling frequency characteristics, these methods have demonstrated the effectiveness of frequency information for alleviating modality discrepancy and improving representation learning.

Despite their effectiveness, existing frequency-based methods primarily focus on frequency enhancement, spectral feature modeling, or frequency-aware modality alignment. Relatively less attention has been paid to how high-frequency representations can be explicitly regularized to simultaneously suppress modality-specific distortions and preserve identity-discriminative patterns. As a result, the potential of high-frequency representations for cross-modal discriminative learning remains insufficiently explored.

\section{Method}
\subsection{Overview}

\begin{figure*}[!t]
\centering
\includegraphics[width=7.0in,height=3.40in]{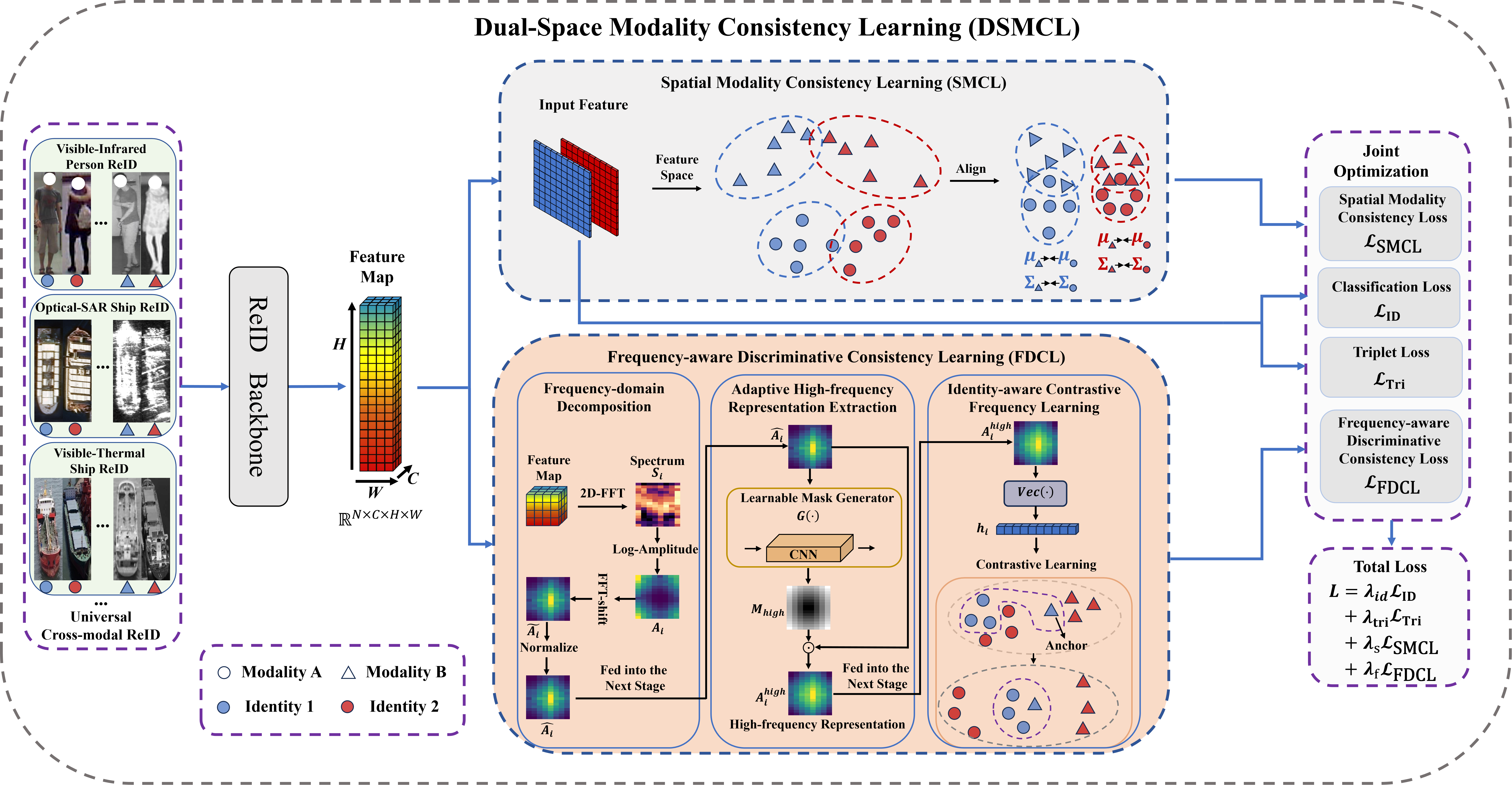}
\caption{Overview of the proposed Dual-Space Modality Consistency Learning (DSMCL) framework. As a unified and plug-and-play modality consistency learning framework, DSMCL can be readily integrated into different backbone architectures and cross-modal scenarios. Given heterogeneous cross-modal inputs, the proposed SMCL performs class-wise spatial distribution alignment across modalities, while FDCL enhances cross-modal discriminability through identity-aware high-frequency contrastive learning. The two complementary objectives are jointly optimized with identity classification and triplet losses for universal cross-modal re-identification.}
\label{fig:framework}
\end{figure*}

Given a cross-modal ReID dataset 
$\mathcal{D}=\{(x_i,y_i,m_i)\}_{i=1}^{N}$,
where $x_i$ denotes an input image, $y_i$ denotes the identity label, and $m_i$ denotes the modality label, the primary objective of cross-modal ReID is formulated as learning a robust feature extractor $f_{\theta}(\cdot)$ that maps heterogeneous modality samples into a unified embedding space. In this space, samples belonging to the same identity should remain compact regardless of modality variations, while samples from different identities should be well separated.

To achieve this objective, we propose a novel Dual-Space Modality Consistency Learning (DSMCL) framework, which comprehensively models modality consistency from both the spatial feature space and the frequency representation space. As illustrated in Fig.~\ref{fig:framework}, the proposed DSMCL consists of two complementary components: Spatial Modality Consistency Learning (SMCL) and Frequency-aware Discriminative Consistency Learning (FDCL).

Specifically, the SMCL branch reduces feature distribution discrepancy across heterogeneous modalities through Gaussian-based class-wise distribution alignment, encouraging same-identity samples from different modalities to remain compact in the embedding space.

Meanwhile, to further enhance robustness, DSMCL introduces the FDCL branch to explicitly model modality discrepancy in high-frequency representations. FDCL transforms intermediate feature maps into the frequency domain and selectively regularizes high-frequency components via identity-aware cross-modal contrastive consistency learning.

Finally, the spatial consistency branch and the frequency-aware consistency branch are jointly optimized together with identity classification and triplet objectives, enabling DSMCL to learn robust and discriminative cross-modal representations for universal cross-modal ReID.

\subsection{Spatial Modality Consistency Learning}

Cross-modal ReID inevitably suffers from significant feature distribution discrepancy across heterogeneous modalities, causing same-identity samples from different modalities to become scattered in the embedding space. To effectively alleviate this issue, we employ a Spatial Modality Consistency Learning (SMCL) strategy to explicitly reduce cross-modal feature distribution discrepancy.

Specifically, given a mini-batch
$\mathcal{B}=\{(x_i,y_i,m_i)\}_{i=1}^{B}$
containing samples from multiple modalities, we first extract spatial features using a backbone network parameterized by $\theta$:
\begin{equation}
f_i=f_{\theta}(x_i),
\end{equation}
where $f_i \in \mathbb{R}^{d}$ denotes the embedding feature of the $i$-th sample, and $\theta$ denotes the learnable parameters of the backbone network.

For each identity class $c$ under modality $m$, we define its feature set as:
\begin{equation}
S_m^c=\{f_i \mid y_i=c,\; m_i=m\}.
\end{equation}

Deep feature representations aggregate responses induced by multiple latent factors. Under the common assumption that these contributions are weakly dependent and have finite variance, the central limit theorem motivates approximating each class-conditional feature distribution by a multivariate Gaussian. This approximation characterizes the central tendency and dispersion of each class-conditional feature distribution through its mean vector and covariance matrix, respectively. Accordingly, we model the feature distribution of identity $c$ under modality $m$ as: 
\begin{equation}
p_m^c(f)=\mathcal{N}\left(\mu_m^c,\Sigma_m^c\right),
\end{equation}
where $\mu_m^c \in \mathbb{R}^{d}$ and
$\Sigma_m^c \in \mathbb{R}^{d \times d}$ denote the class-wise mean vector and covariance matrix, respectively. They are estimated as: 
\begin{equation}
\mu_m^c=
\frac{1}{|S_m^c|}
\sum_{f_i \in S_m^c} f_i,
\end{equation}
and
\begin{equation}
\Sigma_m^c=
\frac{1}{|S_m^c|}
\sum_{f_i \in S_m^c}
\left(f_i-\mu_m^c\right)
\left(f_i-\mu_m^c\right)^{\top}.
\end{equation}

The Gaussian representation characterizes each class-modality distribution through its first- and second-order statistics. We therefore measure the discrepancy between modalities using the Wasserstein-2 distance, which admits a closed-form expression between Gaussian distributions. Given
\begin{equation}
p_{m_a}^c=\mathcal{N}\left(\mu_{m_a}^c,\Sigma_{m_a}^c\right)
\end{equation}
and
\begin{equation}
p_{m_b}^c=\mathcal{N}\left(\mu_{m_b}^c,\Sigma_{m_b}^c\right),
\end{equation}
their squared Wasserstein-2 distance is formulated as:
\begin{equation}
\label{w-2_distance}
\begin{aligned}
\mathcal{W}_2^2
=&\,
\|\mu_{m_a}^c-\mu_{m_b}^c\|_2^2 \\
&+
\mathrm{Tr}
\left(
\Sigma_{m_a}^c
+
\Sigma_{m_b}^c
-
2
\left(
(\Sigma_{m_a}^c)^{\frac{1}{2}}
\Sigma_{m_b}^c
(\Sigma_{m_a}^c)^{\frac{1}{2}}
\right)^{\frac{1}{2}}
\right),
\end{aligned}
\end{equation}
where $\mathrm{Tr}(\cdot)$ denotes the trace operator.

In mini-batch training, the number of class-specific samples
$|S_m^c|$ is typically much smaller than the feature dimension $d$.
Consequently, the empirical covariance matrix satisfies
$\operatorname{rank}(\Sigma_m^c) \leq |S_m^c|-1$
and is generally rank-deficient, making its estimation statistically
unreliable. Moreover, evaluating the Wasserstein-2 distance between full
covariance matrices requires costly and potentially unstable matrix square-root
operations. We therefore retain the dimension-wise variances and approximate
the full covariance matrix by a diagonal matrix:
\begin{equation}
\Sigma_m^c
\approx
\operatorname{Diag}\left((\sigma_m^c)^2\right),
\end{equation}
where $\sigma_m^c \in \mathbb{R}^{d}$ denotes the class-wise standard deviation vector, $(\sigma_m^c)^2$ is computed element-wise, and $\operatorname{Diag}(\cdot)$ constructs a diagonal matrix from its vector argument. This approximation reduces the number of covariance
statistics from $\mathcal{O}(d^2)$ to $\mathcal{O}(d)$ while preserving the
feature dispersion along each dimension. It therefore enables efficient and
stable class-wise distribution alignment within each mini-batch. The diagonal approximation does not assume that the feature dimensions are
strictly independent; instead, it provides a practical trade-off between
distributional expressiveness, estimation reliability, and computational
efficiency.


Under the diagonal covariance assumption, the covariance interaction term in Eq.~(\ref{w-2_distance}) can be simplified into the Euclidean distance between standard deviation vectors. The final spatial modality consistency objective is formulated as:
\begin{equation}
\label{l_smcl}
\mathcal{L}_{\text{SMCL}}
=
\frac{1}{|\mathcal{C}|}
\sum_{c\in\mathcal{C}}
\sum_{(m_a,m_b)}
\left(
\|\mu_{m_a}^c-\mu_{m_b}^c\|_2^2
+
\|\sigma_{m_a}^c-\sigma_{m_b}^c\|_2^2
\right),
\end{equation}
where $\mathcal{C}$ denotes the identity set within the current mini-batch.

By aligning class-wise feature distributions across heterogeneous modalities, SMCL encourages same-identity samples from different modalities to remain compact in the spatial embedding space, thereby reducing modality-induced distribution discrepancy for better generalization.

\subsection{Frequency-Aware Discriminative Consistency Learning}

To further model modality discrepancy in the frequency domain, we propose a Frequency-aware Discriminative Consistency Learning (FDCL) strategy that explicitly regularizes high-frequency representations. Different from conventional spatial feature alignment methods that optimize feature consistency only in the embedding space, FDCL additionally analyzes modality discrepancy from the perspective of frequency-domain representation learning. The key idea of FDCL is that heterogeneous imaging mechanisms predominantly affect high-frequency responses while discriminative identity information is also mainly concentrated in high-frequency components. Therefore, directly aligning full-spectrum representations may suppress discriminative identity details together with modality-specific distortions, leading to over-smoothed feature representations. 

To address this issue, FDCL consists of three stages: frequency-domain feature decomposition, adaptive high-frequency representation extraction, and identity-aware contrastive frequency learning. Specifically, FDCL first transforms spatial representations into the frequency domain to explicitly characterize modality-dependent frequency responses. It then adaptively extracts informative high-frequency components through a learnable frequency mask generator. Finally, identity-aware contrastive optimization is performed in the high-frequency representation space to preserve discriminative cross-modal identity patterns while alleviating modality-specific frequency distortions.

\subsubsection{Frequency-domain Feature Decomposition}

Specifically, given an intermediate feature map extracted from the backbone network:
\begin{equation}
F_i \in \mathbb{R}^{C\times H\times W},
\end{equation}
where $C$, $H$, and $W$ denote the channel number, height, and width, respectively, we first transform the spatial representation into the frequency domain through a two-dimensional discrete Fourier transform (2D-DFT):
\begin{equation}
S_i(c,u,v)
=
\sum_{x=0}^{H-1}
\sum_{y=0}^{W-1}
F_i(c,x,y)
e^{-j2\pi
\left(
\frac{ux}{H}
+
\frac{vy}{W}
\right)},
\end{equation}
where $(x,y)$ and $(u,v)$ denote the spatial coordinates and frequency coordinates, respectively. $S_i(c,u,v)\in\mathbb{C}$ represents the complex-valued frequency-domain representation of the $c$-th channel.

According to the polar representation of complex numbers, each complex-valued frequency coefficient can be represented by its magnitude and phase. Specifically, for a complex coefficient
$S_i(c,u,v)=a+jb$, it can be rewritten as:
\begin{equation}
S_i(c,u,v)
=
|S_i(c,u,v)|e^{j\phi_i(c,u,v)},
\end{equation}
where
\begin{equation}
|S_i(c,u,v)|
=
\sqrt{a^2+b^2},
\end{equation}
and $\phi_i(c,u,v)=\arctan(b/a)$ denotes the phase angle. Therefore, the Fourier coefficient can be naturally decomposed into an amplitude component and a phase component. The amplitude indicates the response strength of a specific frequency component, while the phase mainly preserves spatial structural information and details.

Since heterogeneous imaging mechanisms primarily perturb the frequency response magnitude of different spatial frequencies, modality discrepancy is more directly reflected in the amplitude spectrum. Therefore, we compute the log-amplitude spectrum of the transformed frequency representation:
\begin{equation}
A_i(c,u,v)
=
\log(|S_i(c,u,v)|+\epsilon),
\end{equation}
where $A_i(c,u,v)$ denotes the logarithmic amplitude spectrum at frequency coordinate $(u,v)$ and $\epsilon$ is a small constant for numerical stability. The logarithmic transformation performs dynamic range compression on Fourier amplitudes and avoids the dominance of extremely large low-frequency responses, thereby producing a more balanced frequency magnitude distribution for subsequent high-frequency representation learning and analysis in the frequency domain.

To facilitate explicit frequency decomposition, we further apply
$\mathrm{FFTshift}(\cdot)$
to rearrange the frequency spectrum by moving low-frequency components to the center of the spectrum:
\begin{equation}
\tilde{A}_i
=
\mathrm{FFTshift}(A_i).
\end{equation}

After frequency rearrangement, the shifted spectrum exhibits an explicit
frequency structure, with low-frequency components concentrated near the
spectrum center and high-frequency components distributed in peripheral
regions. However, large variations in spectral magnitude across modalities
and samples may lead to unstable optimization. We therefore standardize the
shifted spectrum as
\begin{equation}
\hat{A}_i
=
\frac{
\tilde{A}_i-\operatorname{Mean}(\tilde{A}_i)
}{
\operatorname{Std}(\tilde{A}_i)+\epsilon
},
\end{equation}
where $\operatorname{Mean}(\cdot)$ and
$\operatorname{Std}(\cdot)$ denote the mean and standard deviation operators,
respectively, and $\epsilon$ is a small constant ensuring numerical stability.

\subsubsection{Adaptive High-frequency Representation Extraction}

Different modalities may exhibit significantly different frequency response distributions. For example, infrared modalities often suppress high-frequency texture responses due to thermal smoothing, while SAR modalities introduce high-frequency perturbations caused by speckle noise. Therefore, manually designed frequency partitions may fail to capture modality-specific discriminative regions.

To address this issue, instead of using a fixed frequency partition strategy, FDCL adopts a lightweight learnable mask generator to adaptively determine discriminative high-frequency mask regions:
\begin{equation}
M_{\text{high}}
=
\operatorname{Sigmoid}\left(G(\hat{A}_i)\right),
\end{equation}
where $G(\cdot)$ denotes a lightweight convolutional mask generator and
$\operatorname{Sigmoid}(\cdot)$ denotes the sigmoid activation function.

Accordingly, the high-frequency representation can be obtained by:
\begin{equation}
A_i^{\text{high}}
=
M_{\text{high}}
\odot
\hat{A}_i,
\end{equation}
where $\odot$ denotes element-wise multiplication.

Compared with manually designed radial frequency decomposition, the learnable mask generator enables FDCL to dynamically focus on informative high-frequency regions that are beneficial for cross-modal identity discrimination. Consequently, the extracted high-frequency representations become more adaptive to heterogeneous modality characteristics.

\subsubsection{Identity-aware Contrastive Frequency Learning}

Based on the extracted high-frequency representations, FDCL further performs identity-aware cross-modal consistency learning in the frequency domain. From the perspective of representation learning, directly minimizing the distance between high-frequency spectra may suppress discriminative identity patterns together with modality-specific distortions, leading to over-smoothed representations with reduced discriminative capability. Therefore, FDCL adopts identity-aware contrastive learning instead of direct frequency distribution alignment.

To facilitate contrastive similarity computation, the extracted high-frequency representation is further flattened into a feature vector:
\begin{equation}
h_i
=
\mathrm{Vec}(A_i^{\text{high}}),
\end{equation}
where $\mathrm{Vec}(\cdot)$ denotes the vectorization operation.

For each anchor sample $i$, we define the positive sample set $P(i)$ as samples sharing the same identity but belonging to different modalities, while the negative sample set $N(i)$ contains samples from different identities. Based on these cross-modal relationships, FDCL performs supervised contrastive optimization in the high-frequency representation space:
\begin{equation}
\label{l_fdcl}
\mathcal{L}_{\text{FDCL}}
=
-
\frac{1}{|\mathcal{V}|}
\sum_{i\in\mathcal{V}}
\log
\frac{
\sum_{p\in P(i)}
\exp(\mathrm{Sim}(h_i,h_p)/\tau)
}{
\sum_{a\in P(i)\cup N(i)}
\exp(\mathrm{Sim}(h_i,h_a)/\tau)
},
\end{equation}
where $\mathrm{Sim}(\cdot,\cdot)$ denotes cosine similarity, $\tau$ is the temperature coefficient, and $\mathcal{V}$ denotes the valid anchor set.

The objective of Eq.~(\ref{l_fdcl}) is to maximize the similarity between cross-modal samples belonging to the same identity while simultaneously separating samples from different identities in the high-frequency representation space. Consequently, the learned frequency representations are encouraged to preserve identity-consistent high-frequency patterns instead of modality-specific frequency distortions.

Unlike direct frequency distribution matching, FDCL does not enforce strict equality between high-frequency spectra. Instead, it performs identity-aware discriminative consistency learning, which alleviates modality-induced high-frequency discrepancy while preserving discriminative identity details. Consequently, FDCL enables the model to learn identity-consistent yet modality-robust high-frequency representations, thereby improving cross-modal discriminative representation learning in the frequency domain.

\subsection{Dual-space Joint Optimization}

The proposed DSMCL framework jointly models modality consistency from both the spatial feature space and the frequency representation space. Specifically, Spatial Modality Consistency Learning (SMCL) reduces global feature distribution discrepancy across heterogeneous modalities, while Frequency-aware Discriminative Consistency Learning (FDCL) further regularizes identity-consistent high-frequency representations in the frequency domain. The two branches provide complementary constraints for universal cross-modal representation learning.

In addition to the proposed modality consistency objectives, DSMCL is further supervised by standard identity classification loss and triplet loss to preserve discriminative identity information in the embedding space. The identity classification loss is formulated as:
\begin{equation}
\mathcal{L}_{\text{ID}}
=
-
\sum_{i=1}^{B}
y_i
\log p_i,
\end{equation}
where $p_i$ denotes the predicted identity probability of the $i$-th sample.

The triplet loss is formulated as:
\begin{equation}
\mathcal{L}_{\text{Tri}}
=
\sum_{i=1}^{B}
\left[
d(f_i,f_i^{p})
-
d(f_i,f_i^{n})
+
\alpha
\right]_+,
\end{equation}
where $f_i^{p}$ and $f_i^{n}$ denote the hardest positive and hardest negative samples of anchor feature $f_i$, respectively, $d(\cdot,\cdot)$ denotes the Euclidean distance, and $\alpha$ is the margin parameter.

Based on the spatial consistency objective in Eq.~(\ref{l_smcl}) and the frequency-aware discriminative objective in Eq.~(\ref{l_fdcl}), the overall optimization objective of DSMCL is formulated as:
\begin{equation}
\mathcal{L}
=
\lambda_{\text{id}}
\mathcal{L}_{\text{ID}}
+
\lambda_{\text{tri}}
\mathcal{L}_{\text{Tri}}
+
\lambda_s
\mathcal{L}_{\text{SMCL}}
+
\lambda_f
\mathcal{L}_{\text{FDCL}},
\end{equation}
where $\lambda_{\text{id}}$ and $\lambda_{\text{tri}}$ denote the balancing coefficients of the identity classification loss and triplet loss, respectively, while $\lambda_s$ and $\lambda_f$ are balancing coefficients for spatial consistency learning and frequency-aware discriminative consistency learning. 

Specifically, $\mathcal{L}_{\text{ID}}$ supervises identity-level semantic discrimination in the embedding space, while $\mathcal{L}_{\text{Tri}}$ further constrains the relative distance relationship among samples by pulling same-identity samples closer and pushing different identities farther apart. Built upon these standard ReID objectives, SMCL explicitly aligns class-wise feature distributions across heterogeneous modalities in the spatial embedding space, whereas FDCL further regularizes modality-robust high-frequency discriminative representations in the frequency domain for robust feature learning.

Consequently, the proposed dual-space optimization framework simultaneously reduces modality-induced distribution discrepancy and preserves discriminative identity-aware frequency patterns. Through the joint optimization of spatial consistency and frequency-aware discriminative consistency, DSMCL learns robust representations that effectively adapt to diverse cross-modal scenarios.

\begin{table*}[!t]
\caption{Comparison between the Proposed DSMCL and Some State-of-the-Art Methods on the SYSU-MM01 Dataset. The Symbol “*” Denotes The Reproduction Results Based on The Official Code. The Symbol ``-'' Indicates That Metrics Were Not Reported in The Source Paper. \textbf{Bold} Values Indicate the Best Performance.\label{tab:SYSU}}
\centering
\setlength{\tabcolsep}{5pt}
\begin{tabular}{c|c|ccc|ccc|ccc|ccc}
\hline 
\multirow{3}{*}{Methods} & \multirow{3}{*}{Venue} & \multicolumn{6}{c|}{All-search}&\multicolumn{6}{c}{Indoor-search}\\
\cline {3-14}
& &\multicolumn{3}{c|}{Single-shot}&\multicolumn{3}{c|}{Multi-shot}&\multicolumn{3}{c|}{Single-shot}&\multicolumn{3}{c}{Multi-shot}\\
\cline {3-14}
& &Rank1&Rank10&mAP&Rank1&Rank10&mAP&Rank1&Rank10&mAP&Rank1&Rank10&mAP\\
\cline{1-14}
AGW \cite{survey_ye} & TPAMI'21 & 47.5 & 84.4 & 47.7 & - & - & - &  54.2 & 91.1 & 63.0 & - & - & - \\
DEEN \cite{deen} & CVPR'23 & 74.7 & 97.6 & 71.8 & - & - & - & 80.3 & 99.0 & 83.3 & - & - & - \\
PartMix \cite{partmix} & CVPR'23 & 77.8 & - & 74.6 & 80.5 & - & 69.8 & 81.5 & - & 84.4 & 88.0 & - & 89.0 \\
MUN \cite{MUN} & ICCV'23 & 76.2 & 97.8 & 73.8 & - & - & - & 79.4 & 98.1 & 82.1 & - & - & -\\
CAJ$_+$ \cite{caj+} & TPAMI'23 & 71.5 & 96.2 & 68.2 & - & - & - & 78.4 & 98.4 & 82.0 & - & - & -  \\
MV$^2$D \cite{mv2d} & TPAMI'23 & 71.7 & 96.3 & 68.0 & - & - & - & 79.1 & 99.0 & 81.8 & - & - & - \\
HOS-Net \cite{HOSnet} & AAAI'24 & 75.6 & - & 74.2 & - & - & - & 84.2 & - & 86.7 & - & - & - \\
WRIM-Net \cite{wrim} & ECCV'24 & 77.4 & - & 75.4 & 83.2 & - & 71.1 & 86.2 & - & 88.1 & 92.1 & - & 84.6 \\
DNS \cite{dns} & ECCV'24 & 72.3 & - & 74.4 & - & - & - & 84.2 & - & 86.8 & - & - & - \\
RLE \cite{rle} & NIPS'24 & 75.4 & 97.7 & 72.4 & - & - & - & 84.7 & 99.3 & 87.0 & - & - & - \\
CIIM \cite{ciim} & TPAMI'24 & 57.2 & - & 56.1 & - & - & - & 61.7 & - & 69.8 & - & - & - \\
MSSF \cite{mssf} & TMM'24 & 70.6 & 96.2 & 67.5 & - & - & - & 76.0 & 98.1 & 80.2 & - & - & - \\
DMA \cite{dma} & TIFS'24 & 74.6 & - & 70.4 & 83.0 & - & 65.6 & 82.9 & - &  85.1 & 91.3 & - & 80.5 \\
DiVE \cite{dive} & AAAI'25 & 79.1 & - & 74.9 & - & - & - & 83.0 & - & 85.9 & - & - & -\\ 
AGPI$^2$ \cite{AGPI} & TIFS'25 & 72.2 & 97.0 & 70.6 & - & - & - & 83.5 & 98.6 & 84.3 & - & - & -\\
DSFAD \cite{DSFAD} & TIFS'25 & 78.4 & 97.5 & 74.7 & 84.6 & 98.5 & 69.9 & 85.1 & 99.2 & 87.2 & 91.5 & 99.7 & 82.7 \\
STAR-ReID \cite{STAR-REID} & TIFS'25 & 82.9 & - & 80.5 & - & - & - & 88.0 & - & 89.6 & - & - & -\\
FDNM \cite{fdnm} & TIFS'25 & 77.8 & 97.8 & 75.1 & 82.7 & 98.9 & 70.9 & 87.3 & 99.4 & 89.1 & 92.2 & 99.8 & 83.8 \\
CSDN \cite{csdn} & TMM'25 & 76.7 & 97.2 & 73.0 & 83.5 & 98.8 & 67.9 & 84.5 & 99.2 & 86.8 & 91.3 & 99.8 & 82.2 \\
DSAF \cite{dsaf} & TMM'25 & 76.7 & 97.7 & 73.2 & - & - & - & 83.5 & 99.3 & 83.8 & - & - & - \\ 
HTCR \cite{htcr} & TMM'25 & 74.6 & - & 72.1 & - & - & - & 83.0 & - & 85.6 & - & - & -\\
HOH-Net \cite{HOH-Net} & TCSVT'25 & 76.2 & - & 74.5 & - & - & - & 84.4 & - & 87.2  & - & - & -  \\
DMPF \cite{dmpf} & TNNLS'25 & 76.4 & 96.4 & 71.6 & - & - & - & 82.3 & 98.6 & 84.9 &  - & - & - \\ 
CycleTrans \cite{Cycletrans} & TNNLS'25 & 76.6 & 97.2 & 72.6 & 82.8 & 98.6 & 68.5 & 87.2 & 99.6 & 84.9 & 91.2 & 99.8 & 81.4 \\
MHN \cite{mhn} & INFFUS'25 & 75.1 &  97.3 & 71.8 & - & - & - & 81.6 & 98.9 & 85.1 &  - & - & -\\
MFEN \cite{mfen} & CVPR'26 & 80.9 & 97.1 & 76.6 & - & - & - & 87.9 & 99.4 & 88.1 & - & - & - \\
CMOI \cite{cmoi_cvpr26} & CVPR'26 & 75.6 & - & 73.2 & - & - & - & 84.6 & - & 86.3 & - & - & - \\
ICSD \cite{icsd} & TIP'26 & 77.9 & 98.5 & 74.7 & - & - & - & 85.7 & \textbf{99.7} & 87.8 & - & - & -  \\ 
CMSD \cite{cmsd} & TGRS'26 & 77.7 & 98.1 & 74.2 & - & - & - & 86.3 & 99.3 & 88.6 & - & - & -\\
FFD-Net \cite{FFD-net} & EAAI'26 & 73.7 & 96.8 & 68.2 & - & - & - & 75.6 & 97.1 & 78.5 & - & - & - \\

\cline{1-14}
IDKL* \cite{idkl} & CVPR'24 & 80.9 & 97.2 & 78.5 & 83.5 & 98.4 & 74.9 & 88.2 & 98.5 & 89.6 & 92.2 & 99.7 & 86.4\\
\rowcolor{cat4} IDKL + DSMCL & This paper & \textbf{83.6} & 97.7 & \textbf{81.4} & \textbf{86.1} & 99.1 & \textbf{77.8} & \textbf{92.3} & 99.2 & \textbf{92.2} & \textbf{95.2} & 99.7 & 88.9 \\ 
\cline{1-14}
MFENet* \cite{MFENet} & TCSVT'25 & 77.7 & 97.8 & 75.9 & 82.1 & 98.2 & 71.7 & 87.2 & 98.4 & 89.2 & 92.7 & 99.7 & 86.6\\
\rowcolor{cat4} MFENet + DSMCL  & This paper & 79.6 & \textbf{99.0} & 78.9 & 84.6 & 99.3  & 74.2 & 90.4 & 99.5 & 91.6 & 94.9 & \textbf{100.0} & \textbf{89.2} \\
\cline{1-14}
HSFLNet* \cite{hsflnet} & EAAI'25 & 76.7 & 96.1 & 70.0 & 81.9 & 98.3 & 71.8 & 80.9 & 98.9 & 83.3 & 89.7 & 99.5 & 77.7\\
\rowcolor{cat4} HSFLNet + DSMCL  & This paper & 79.2 & 97.5 & 74.6 & 83.7 & \textbf{99.6} & 73.1 & 83.4 & \textbf{99.7} & 85.7 & 91.7 & 99.8 & 80.3\\
\hline
\end{tabular}
\end{table*}

\section{Experiments}

\begin{table*}[!t]
\caption{Comparison between the Proposed DSMCL and Some State-of-the-Art Methods on the RegDB and LLCM Dataset. The Symbol “*” Denotes The Reproduction Results Based on The Official Code. The Symbol ``-'' Indicates That Metrics Were Not Reported in The Source Paper. \textbf{Bold} Values Indicate the Best Performance.\label{tab:RegDB and LLCM}}
\centering
\setlength{\tabcolsep}{5pt}
\begin{tabular}{c|c|ccc|ccc|ccc|ccc}
\hline 
\multirow{3}{*}{Methods} & \multirow{3}{*}{Venue} & \multicolumn{6}{c|}{RegDB}&\multicolumn{6}{c}{LLCM}\\
\cline {3-14}
& &\multicolumn{3}{c|}{Visible to Infrared}&\multicolumn{3}{c|}{Infrared to Visible}&\multicolumn{3}{c|}{Visible to Infrared}&\multicolumn{3}{c}{Infrared to Visible}\\
\cline {3-14}
& &Rank1&Rank10&mAP&Rank1&Rank10&mAP&Rank1&Rank10&mAP&Rank1&Rank10&mAP\\
\cline{1-14}
AGW \cite{survey_ye} & TPAMI'21 & 70.1 & 86.2 & 66.4 & 70.5 & 87.1 & 65.9 & 51.5 & 81.5 & 55.3 & 40.3 & 71.4 & 48.4 \\
DEEN \cite{deen} & CVPR'23 & 91.1 & 97.8 & 85.1 & 89.5 & 96.8 & 83.4 & 62.5 & 90.3 & 65.8 & 54.9 & 84.9 & 62.9 \\
PartMix \cite{partmix} & CVPR'23 & 85.7 & - & 82.3 & 84.9 & - & 82.5 & - & - & - & - & - & - \\
MUN \cite{MUN} & ICCV'23 & 95.2 & 98.9 & 87.2 & 91.9 & 98.0 & 85.0 & - & - & - & - & - & - \\
CAJ$_+$ \cite{caj+} & TPAMI'23 & 85.7 & 95.5 & 79.7 & 84.9 & 95.9 & 78.6 & - & - & - & - & - & - \\
MV$^2$D \cite{mv2d} & TPAMI'23 & 83.1 & - & 80.1 & 81.2 & - & 78.4 & - & - & - & - & - & - \\
HOS-Net \cite{HOSnet} & AAAI'24 & 94.7 & - & 90.4 & 93.3 & - & 89.2 & 64.9 & - & 67.9 & 56.4 & - & 63.2 \\
WRIM-Net \cite{wrim} & ECCV'24 & 94.5 & - & 90.5 & 93.7 & - & 89.7 & 67.0 & - & 69.2 & 58.4 & - & 64.8 \\
DNS \cite{dns} & ECCV'24 & 93.0 & - & 88.6 & 93.5 & - & 88.1 & 66.0 & - & 68.6 & 57.5 & - & 64.1 \\
RLE \cite{rle} & NIPS'24 & 92.8 & 97.9 & 88.6 & 91.0 & 97.5 & 86.6 & - & - & - & - & - & - \\
CIIM \cite{ciim} & TPAMI'24 & 76.5 & - & 71.0 & 74.8 & - & 69.7 &  - & - & - & - & - & - \\
DMA \cite{dma} & TIFS'24 & 93.3 & - &  83.3 & 91.5 & - & 86.8 & - & - & - & - & - & - \\
MSSF \cite{mssf} & TMM'24 & 85.3 & - & 76.4 & 83.9 & - & 75.2 & - & - & - & - & - & - \\
MDANet \cite{mdanet} & TMM'24 & 94.0 & - & 83.7 & 93.7 & - & 82.5 & - & - & - & - & - & - \\
AGPI$^2$ \cite{AGPI} & TIFS'25 & 89.0 & 98.2 & 83.9 & 87.9 & 97.2 & 83.0 & 61.8 & - & 65.1 & 56.5 & - & 62.8 \\
DSFAD \cite{DSFAD} & TIFS'25 & 95.8 & 99.6 & 90.2 & 94.3 & 99.2 & 89.4 & 66.2 & 91.5 & 68.9 & 57.5 & 86.2 & 64.1 \\
STAR-ReID \cite{STAR-REID} & TIFS'25 & 91.9 & - & 93.3 & 90.5 & - & 91.8 & - & - & - & - & - & -\\
FDNM \cite{fdnm} & TIFS'25 & 95.5 & 99.0 & 90.0 & 94.0 & 98.5 & 88.7 & 70.2 & 91.1 & 55.8 & 56.6 & 84.1 & 62.7 \\
CSDN \cite{csdn} & TMM'25 & 95.4 & 98.8 & 87.7 & 92.3 & 98.0 & 85.5 & 63.7 & 90.9 & 66.5 & 55.8 & 85.6 & 63.5\\
DSAF \cite{dsaf} & TMM'25 & 93.3 & 99.0 & 87.2 & 92.6 & 98.2 & 86.4 & 65.4 & - & 68.2 & 57.3 & - & 64.3\\
HTCR \cite{htcr} & TMM'25 & 92.6 & - & 87.9 & 90.2 & - & 85.6 & 63.9 & - & 66.6 & 55.4 & - & 62.0 \\
HOH-Net \cite{HOH-Net} & TCSVT'25 & 95.1 & - & 90.7 & 93.7 & - & 89.5 & 65.7 & - & 68.3 & 56.8 & - & 63.5 \\
DMPF \cite{dmpf} & TNNLS'25 & 88.8 & 97.4 & 81.0 & 88.9 & 97.6 & 81.9 &  - & - & - & - & - & -\\
CycleTrans \cite{Cycletrans} & TNNLS'25 & 91.5 & - &  87.0 & 91.3 & - & 84.9 &  - & - & - & - & - & - \\
MHN \cite{mhn} & INFFUS'25 & 94.0 & 98.6 & 88.4 & 93.0 & 97.4 & 86.5 & 56.7 & 83.7 & 64.5 & 69.8 & 91.9 & 55.2 \\
MFEN \cite{mfen} & CVPR'26 & 94.9 & 99.1 & 90.1 & 94.1 & 98.8 & 90.3 & 67.9 & 91.0 & 69.8 & 59.0 & 85.6 & 65.3 \\
BIT \cite{bit} & CVPR'26 & 96.1 & - & 92.4 & 93.7 & - & 93.5 & 73.1 & - & 69.2 & 66.7 & - & 67.2\\
CMOI \cite{cmoi_cvpr26} & CVPR'26 & 93.5 & - & 85.3 & 92.4 & - & 83.8 & 70.8 & - & 55.9 & 58.5 & - & 64.5 \\
ICSD \cite{icsd} & TIP'26 & 95.9 & 98.5 & 92.2 &  95.4 & 98.6 & 91.0 & 65.8 & 92.0 & 68.4 & 58.4 & 85.3 & 64.8 \\
CMSD \cite{cmsd} & TGRS'26 & 94.8 & 98.5 & 90.7 & 94.4 & 98.5 & 90.7 & 66.7 & 91.7 & 69.1 & 57.7 & 85.1 & 64.1 \\
FFD-Net \cite{FFD-net} & EAAI'26 & 95.6 & 99.0 & 88.2 & 93.6 & 99.1 & 86.0 & 70.4 & 86.2 & 52.0 & 62.8 & 79.6 & 45.9 \\

\cline{1-14}
IDKL* \cite{idkl} & CVPR'24 & 93.6 & 99.8 & 88.7 & 93.2 & 99.5 & 87.6 & 72.9 & 94.7 & 62.3 & 70.0 & 90.3 & 61.3\\
\rowcolor{cat4} IDKL + DSMCL & This paper & 95.4 & \textbf{100.0} & 90.1 & 95.8 & \textbf{100.0} & 89.5 & \textbf{74.8} & \textbf{95.5} & 64.2 & \textbf{71.4} & \textbf{92.2} & 63.9\\
\cline{1-14} 

MFENet* \cite{MFENet} & TCSVT'25 & 88.4 & 96.5 & 83.3 & 90.5 & 97.5 & 84.4 & 69.2 & 92.8 & 71.2 & 61.2 & 87.8 & 67.6\\
\rowcolor{cat4} MFENet + DSMCL & This paper & 90.5 & 97.7 & 84.9 & 91.7 & 97.6 & 84.7 & 71.6 & 94.0 & \textbf{73.7} & 63.7 & 90.2 & \textbf{69.8}\\
\cline{1-14}
HSFLNet* \cite{hsflnet} & EAAI'25 & 96.1 & 98.8 & 92.3 & 92.2 & 96.8 & 88.7 & 56.6 & 84.5 & 58.5 & 48.8 & 77.2 & 54.8\\
\rowcolor{cat4} HSFLNet + DSMCL  & This paper & \textbf{97.3} & 99.5 & \textbf{93.7} & \textbf{96.1} & 99.2 & \textbf{93.9} & 60.9 & 88.2 & 62.5 & 51.7 & 81.1 & 57.8 \\
\end{tabular}
\end{table*}

\subsection{Experimental Setup}
\subsubsection{Datasets and Evaluation Metrics} We evaluate the proposed DSMCL method on five cross-modal ReID benchmarks, including three visible-infrared person ReID datasets: SYSU-MM01 \cite{sysumm01}, RegDB \cite{regdb}, and LLCM \cite{deen}, as well as two cross-modal ship ReID datasets: HOSS-ReID \cite{transoss} and CMShipReID \cite{cmshipreid}. These datasets cover both person and ship retrieval scenarios with diverse modality combinations, enabling comprehensive evaluation of the proposed universal cross-modal representation learning framework.

\textbf{SYSU-MM01} \cite{sysumm01} is a widely used benchmark for visible-infrared person ReID, containing 30,071 visible images and 15,792 infrared images captured by 4 RGB cameras and 2 infrared cameras. The dataset includes 491 identities, where 22,258 visible images and 11,909 infrared images from 395 identities are used for training, while the remaining identities are reserved for testing. The dataset provides four evaluation protocols, including all-search and indoor-search settings, each evaluated under single-shot and multi-shot gallery modes, where infrared images are used as queries and visible images are used as galleries.

\textbf{RegDB} \cite{regdb} is a common benchmark for visible-infrared person ReID, containing 412 identities captured by paired visible and infrared cameras. Each identity includes 10 visible images and 10 infrared images, resulting in 4,120 visible images and 4,120 infrared images in total. The dataset is evenly split into 206 identities for training and 206 identities for testing. Following the standard protocol, two evaluation settings are provided: Visible-to-Infrared and Infrared-to-Visible.

\textbf{LLCM} \cite{deen} is a large-scale benchmark for visible-infrared person ReID under low-light conditions, containing 46,767 visible and infrared images of 1,064 identities captured by 18 cameras across diverse indoor and outdoor environments. Following the standard protocol, 713 identities with 16,946 visible images and 13,975 infrared images are used for training, while the remaining 351 identities are used for testing. The dataset is evaluated under both Visible-to-Infrared and Infrared-to-Visible retrieval settings.

\textbf{HOSS-ReID} \cite{transoss} is the first benchmark for optical-SAR cross-modal ship ReID, containing 1,832 ship images from 449 identities, including 1,065 optical images and 767 SAR images. Following the standard protocol, 361 identities with 574 optical images and 489 SAR images are used for training, while the remaining identities are used for testing. The dataset provides three evaluation settings: ALL-to-ALL, Optical-to-SAR, and SAR-to-Optical retrieval.

\textbf{CMShipReID} \cite{cmshipreid} is a cross-modal ship ReID benchmark containing visible (VIS), near-infrared (NIR), and thermal-infrared (TIR) ship images collected by unmanned aerial vehicle platforms. The dataset consists of 7,927 images from 138 ship identities, including 3,071 VIS images, 2,530 NIR images, and 2,326 TIR images. Following the standard protocol, 100 identities are used for training, while the remaining 38 identities are used for testing. The dataset provides six cross-modal retrieval settings covering all possible bidirectional modality combinations, including VIS-to-NIR, NIR-to-VIS, VIS-to-TIR, TIR-to-VIS, NIR-to-TIR, and TIR-to-NIR.

For the visible-infrared person ReID benchmarks, including SYSU-MM01 \cite{sysumm01}, RegDB \cite{regdb}, and LLCM \cite{deen}, we adopt Rank1, Rank10, and mean Average Precision (mAP) as the evaluation metrics. For the cross-modal ship ReID benchmarks, including HOSS-ReID \cite{transoss} and CMShipReID \cite{cmshipreid}, we report Rank1, Rank5, Rank10, and mAP for performance evaluation on these benchmarks.

\subsection{Implementation Details}


DSMCL is a plug-and-play framework that can be integrated into existing cross-modal ReID models. To evaluate its effectiveness and generalization capability, DSMCL is incorporated into representative baselines across different cross-modal scenarios. Specifically, for visible-infrared person ReID benchmarks, we adopt IDKL \cite{idkl}, MFENet \cite{MFENet}, and HSFLNet \cite{hsflnet} as baselines. For optical-SAR ship ReID, DSMCL is integrated into IDKL \cite{idkl}, TransOSS \cite{transoss}, and MOS \cite{mos}. For the CMShipReID benchmark, we evaluate DSMCL on TransReID \cite{TransReID}, IDKL \cite{idkl}, MFENet \cite{MFENet}, and HSFLNet \cite{hsflnet}.

Unless otherwise specified, all training configurations follow the official implementations of the corresponding baselines to ensure fair comparisons. The proposed SMCL and FDCL modules are jointly optimized together with the original objectives of each baseline. The balancing coefficients $\lambda_{\text{id}}$ and $\lambda_{\text{tri}}$ follow the official settings of the corresponding baseline models, while $\lambda_s$ and $\lambda_f$ are determined based on validation performance. The temperature coefficient in FDCL is fixed to $\tau=0.2$ for all experiments. This value is chosen empirically and kept constant to maintain consistent scaling of the logits across different datasets.

To stabilize training, a progressive optimization strategy is adopted for the proposed consistency objectives. Specifically, in the spatial domain, SMCL employs a warm-up schedule and is activated after the first 20 epochs. Meanwhile, FDCL adopts a more conservative optimization strategy, remaining inactive during the first 40 epochs and then gradually increasing to its target weight through a linear ramp-up schedule over the subsequent 40 epochs. Such a progressive learning scheme allows the backbone network to first learn reliable identity-discriminative representations before introducing spatial distribution alignment and frequency-domain consistency constraints, thereby improving optimization stability and convergence. All experiments are implemented in PyTorch and conducted on a single NVIDIA A100-SXM4-80GB GPU.

\begin{table*}[!t]
\caption{Comparison between the Proposed DSMCL and Some State-of-the-Art Methods on the HOSS-ReID Dataset. The Symbol “*” Denotes The Reproduction Results Based on The Official Code. The Symbol ``-'' Indicates That Metrics Were Not Reported in The Source Paper. \textbf{Bold} Values Indicate the Best Performance.\label{tab:HOSS-ReID}}
\centering
\setlength{\tabcolsep}{5pt}
\begin{tabular}{c|c|cccc|cccc|cccc}
\hline 
\multirow{2}{*}{Methods} & \multirow{2}{*}{Venue} & \multicolumn{4}{c|}{All to ALL}&\multicolumn{4}{c|}{Optical to SAR}&\multicolumn{4}{c}{SAR to Optical}\\
\cline {3-14}

& &Rank1&Rank5&Rank10&mAP&Rank1&Rank5&Rank10&mAP&Rank1&Rank5&Rank10&mAP\\
\hline
AGW \cite{survey_ye} & TPAMI'21 & 57.4 & 64.2 & 68.8 & 43.6 & 7.7 & 29.2 & 38.5 & 17.2 & 14.9 & 34.3 & 46.3 & 21.1 \\
TransReID \cite{TransReID} & ICCV'21 &  60.8 & 69.3 & 73.9 & 48.1 & 18.5 & 40.0 & 58.5 &  27.3 & 11.9 & 34.3 & 43.3 & 20.9 \\
SOLIDER \cite{SOLIDER} & CVPR'23 & 50.6 & 63.1 & 69.9 & 38.2 & 12.3 & 38.5 & 52.3 & 23.1 & 10.4 & 16.4 & 31.3 & 14.6 \\
DEEN \cite{deen} & CVPR'23 & 58.5 & 64.2 & 66.5 & 43.8 & 21.5 & 44.6 & 60.0 & 31.3 & 22.4 & 40.3 & 53.7 & 27.4 \\
VersReID \cite{versReID} & TPAMI'24 & 59.7 & 70.5 & 78.4 & 49.3 & 13.8 & 40.0 & 61.5 & 25.7 & 17.9 &44.8 & 61.2 & 27.7 \\
MCJA \cite{MCJA} & TCSVT'24 & 59.1 & 67.9 & 73.0 & 47.1 & 10.8 & 27.7 & 38.5 & 18.6 & 14.9 & 28.3 & 43.3 & 19.7 \\
D2InterNet \cite{D2InterNet} & SIGIR'25 & 59.1 & 71.6 & 79.0 & 50.2 & 21.5 & 41.5 & 69.8 & 33.0 & 25.4 & 38.8 & 50.7 & 28.8 \\
\hline
IDKL* \cite{idkl} & CVPR'24 & 64.2 & 76.1 & 84.1 & 55.1 & 36.9 & 64.6 & 75.4 & 51.2 & 31.3 & 61.2 & 74.6 & 35.6\\
\rowcolor{cat4} IDKL + DSMCL  & This paper & 66.5 & 78.4 & 85.2 & 56.7 & 40.0 & 67.7 & 76.9 & 52.4 & 38.8 & 68.7 & 79.1 & 40.6 \\
\hline
TransOSS* \cite{transoss} & ICCV'25 & 66.5 & 77.3 & 83.5 & 56.1 & 35.4 & 58.5 & 73.8 & 47.4 & 31.3 & 56.7 & 71.6 & 38.4\\
\rowcolor{cat4} TransOSS + DSMCL  & This paper & 68.2 & 79.0 & 84.7 & 58.5 & 40.0 & 63.1 & \textbf{78.5} & 50.3 & 38.8 & 64.2 & 77.6 & 42.2\\
\hline
MOS* \cite{mos} & CVPR'26 & 68.8 & 83.5 & 88.6 & 60.4 & 40.0 & 70.8 & 75.4 & 51.4 & 46.3 & 68.7 & 79.1 & 48.7 \\
\rowcolor{cat4} MOS + DSMCL  & This paper & \textbf{70.4} & \textbf{84.7} & \textbf{89.8} & \textbf{62.1} & \textbf{41.5} & \textbf{73.8} & 76.9 & \textbf{52.8} & \textbf{47.8} & \textbf{70.1} & \textbf{82.1} & \textbf{49.8}\\
\hline
\end{tabular}
\end{table*}

\begin{table*}[!t]
\caption{Comparison between the Proposed DSMCL and Some State-of-the-Art Methods on the CMShipReID Dataset under TIR To VIS, VIS To TIR, and NIR To VIS Evaluation Protocols. The Symbol “*” Denotes The Reproduction Results Based on The Official Code. The Symbol ``-'' Indicates That Metrics Were Not Reported in The Source Paper. \textbf{Bold} Values Indicate the Best Performance.\label{tab:CMShip1}}
\centering
\setlength{\tabcolsep}{5pt}
\begin{tabular}{c|c|cccc|cccc|cccc}
\hline 
\multirow{2}{*}{Methods} & \multirow{2}{*}{Venue} & \multicolumn{4}{c|}{TIR to VIS}&\multicolumn{4}{c|}{VIS to TIR} & \multicolumn{4}{c}{NIR to VIS}\\
\cline{3-14}
&&Rank1&Rank5&Rank10&mAP&Rank1&Rank5&Rank10&mAP&Rank1&Rank5&Rank10&mAP\\
\hline
TransReID* \cite{TransReID} & ICCV'21 & 95.6 & 97.9 & 98.9 & 66.8 & 91.4 & 96.0 & 97.9 & 73.4 & 76.9 & 91.7 & 95.7 & 57.0\\
\rowcolor{cat4} TransReID + DSMCL & This paper & 96.6 & 98.2 & 99.0 & 73.9 & \textbf{94.0} & \textbf{97.2} & 98.1 & \textbf{75.4} & \textbf{83.8} & \textbf{94.6} & \textbf{97.3} & \textbf{63.2} \\
\hline
IDKL* \cite{idkl} & CVPR'24 & 94.4 & 98.2 & 99.5 & 55.7 & 82.5 & 95.9 & 99.1 & 52.6 & 68.2 & 89.0 & 95.0 & 41.1\\
\rowcolor{cat4} IDKL + DSMCL & This paper & 95.8 & 98.9 & 99.7 & 58.5 & 85.0 & 96.7 & \textbf{99.4} & 54.5 & 71.7 & 91.5 & 95.4 & 44.7\\
\hline
MFENet* \cite{MFENet} & TCSVT'25 & 96.9 & 98.5 & 99.0 & 73.1 & 90.6 & 95.0 & 96.3 & 72.0 & 79.5 & 91.6 & 94.7 & 60.7\\
\rowcolor{cat4} MFENet + DSMCL & This paper & \textbf{98.7} & \textbf{99.7} & \textbf{99.8} & \textbf{74.3} & 93.7 & 96.9 & 97.7 & 74.4 & 82.9 & 91.5 & 94.9 & 62.5\\
\hline
HSFLNet* \cite{hsflnet} & EAAI'25 & 97.9 & 98.9 & 99.4 & 64.0 & 90.0 & 95.2 & 96.9 & 62.9 & 80.6 & 93.6 & 96.4 & 46.1 \\
\rowcolor{cat4} HSFLNet + DSMCL & This paper & 98.4 & 99.0 & 99.5 & 65.7 & 91.8 & 95.8 & 97.3 & 64.5 & 83.2 & 94.3 & 96.7 & 49.5 \\
\hline
\end{tabular}
\end{table*}

\begin{table*}[!t]
\caption{Comparison between the Proposed DSMCL and Some State-of-the-Art Methods on the CMShipReID Dataset under VIS To NIR, NIR To TIR, and TIR To NIR Evaluation Protocols. The Symbol “*” Denotes The Reproduction Results Based on The Official Code. The Symbol ``-'' Indicates That Metrics Were Not Reported in The Source Paper. \textbf{Bold} Values Indicate the Best Performance.\label{tab:CMShip2}}
\centering
\setlength{\tabcolsep}{5pt}
\begin{tabular}{c|c|cccc|cccc|cccc}
\hline 
\multirow{2}{*}{Methods} & \multirow{2}{*}{Venue} & \multicolumn{4}{c|}{VIS to NIR}& \multicolumn{4}{c|}{NIR to TIR}&\multicolumn{4}{c}{TIR to NIR}\\
\cline{3-14}
&&Rank1&Rank5&Rank10&mAP&Rank1&Rank5&Rank10&mAP&Rank1&Rank5&Rank10&mAP\\
\hline
TransReID* \cite{TransReID} & ICCV'21 & 72.7 & 86.0 & 91.2 & 57.2 & 75.1 & 89.0 & 92.6 & 59.2 & 73.2 & 89.2 & 95.0 & 54.4\\
\rowcolor{cat4} TransReID + DSMCL & This paper & \textbf{79.8} & \textbf{92.8} & 95.7 & \textbf{60.9} & 78.1 & 90.3 & 93.9 & 61.4 & \textbf{77.4} & \textbf{92.1} & \textbf{95.3} & \textbf{60.2}\\
\hline
IDKL* \cite{idkl} & CVPR'24 & 59.8 & 87.9 & 95.3 & 35.7 & 65.6 & 90.6 & 96.4 & 38.6 & 65.8 & 89.8 & 94.8 & 37.9\\
\rowcolor{cat4} IDKL + DSMCL & This paper & 62.3 & 88.1 & \textbf{95.8} & 39.6 & 68.5 & 91.7 & \textbf{96.9} & 41.4 & 66.1 & 90.8 & 95.0 & 38.0\\
\hline
MFENet* \cite{MFENet} & TCSVT'25 & 72.0 & 85.1 & 89.8 & 56.8 & 77.0 & 90.3 & 92.6 & 64.4 & 72.4 & 89.4 & 93.7 & 58.7\\
\rowcolor{cat4} MFENet + DSMCL & This paper & 74.1 & 85.3 & 90.0 & 57.9 & \textbf{81.4} & \textbf{94.7} & 95.1 & \textbf{66.5} & 76.0 & 91.8 & 94.4 & \textbf{60.2}\\
\hline
HSFLNet* \cite{hsflnet} & EAAI'25 & 71.2 & 86.0 & 91.4 & 40.6 & 60.1 & 76.5 & 83.7 & 32.7 & 66.6 & 83.7 & 89.7 & 33.0 \\
\rowcolor{cat4} HSFLNet + DSMCL & This paper & 73.8 & 87.6 & 92.1 & 42.5 & 63.4 & 81.1 & 88.1 & 35.2 & 69.4 & 86.0 & 91.8 & 35.2\\
\hline
\end{tabular}
\end{table*}

\subsection{Comparison with State-of-the-Art Methods}

\subsubsection{Results on Visible-Infrared Person ReID Benchmarks}

Tables~\ref{tab:SYSU} and~\ref{tab:RegDB and LLCM} report the comparison results on SYSU-MM01, RegDB, and LLCM. Overall, DSMCL consistently improves all baseline models across different datasets and evaluation protocols, demonstrating its effectiveness and strong generalization capability for visible-infrared person ReID.

On SYSU-MM01, DSMCL brings consistent performance gains when integrated into different baseline architectures. For example, IDKL+DSMCL improves the Rank-1 accuracy from 80.9\% to 83.6\% and the mAP from 78.5\% to 81.4\% under the All-search Single-shot protocol. Similar improvements are also observed on MFENet and HSFLNet under both All-search and Indoor-search settings. Notably, IDKL+DSMCL achieves the best Rank-1 accuracy of 92.3\% and 95.2\% under Indoor-search Single-shot and Multi-shot protocols, respectively.

On RegDB and LLCM, DSMCL also consistently improves all baseline models under different retrieval protocols. In particular, HSFLNet+DSMCL achieves the best overall performance on RegDB with 97.3\% Rank-1 and 93.7\% mAP under the Visible-to-Infrared protocol. Consistent gains are further observed on the more challenging LLCM benchmark, indicating that DSMCL remains effective under severe modality discrepancies and low-light conditions, which further validates its robustness. Overall, the results on three visible-infrared person ReID benchmarks demonstrate that jointly modeling spatial consistency and frequency-aware discriminative consistency leads to more robust cross-modal representations.

\begin{table*}[!t]
\caption{Component-wise Ablation Study of the Proposed DSMCL on the SYSU-MM01 and RegDB Datasets Under All Evaluation Protocols. \textbf{Bold} Values Indicate the Best Performance.\label{tab:ablation_sysu_and_regcb}}
\centering
\begin{tabular}{c|cc|cc|cc|cc|cc|cc|cc}
\hline 
\multirow{4}{*}{Method} & \multirow{4}{*}{SMCL} & \multirow{4}{*}{FDCL} & \multicolumn{8}{c|}{SYSU-MM01} & \multicolumn{4}{c}{RegDB} \\
\cline {4-15}
 & & & \multicolumn{4}{c|}{All-search} & \multicolumn{4}{c|}{Indoor-search} &  \multicolumn{2}{c|}{\multirow{2}{*}{Visible to Infrared}} & \multicolumn{2}{c}{\multirow{2}{*}{Infrared to Visible}} \\
\cline {4-11}
& & &\multicolumn{2}{c|}{Single-shot}&\multicolumn{2}{c|}{Multi-shot}&\multicolumn{2}{c|}{Single-shot} & \multicolumn{2}{c|}{Multi-shot} &  & \\
\cline {4-15}
& & & Rank1&mAP&Rank1&mAP&Rank1&mAP&Rank1&mAP&Rank1&mAP&Rank1&mAP\\
\hline
IDKL & \ding{55} & \ding{55} & 80.9 & 78.5 & 83.5 & 74.9 & 88.2 & 89.6 & 92.2 & 86.4 & 93.6 & 88.7 & 93.2 & 87.6 \\
IDKL + SMCL & \ding{51} & \ding{55} & 81.5 & 79.6 & 84.8 & 75.3 & 90.7 & 90.9 & 92.7 & 87.3 & 94.1 & 89.5 & 93.9 & 88.1\\
IDKL + FDCL & \ding{55} & \ding{51} & 82.9 & 79.8 & 85.4 & 75.2 & 89.6 & 91.4 & 94.3 & 88.0 & 95.0 & 89.4 & 95.2 & 88.6\\
IDKL + DSMCL & \ding{51} & \ding{51} & \textbf{83.6} & \textbf{81.4} & \textbf{86.1} & \textbf{77.8} & \textbf{92.3} & \textbf{92.2} &  \textbf{95.2} & \textbf{88.9} & \textbf{95.4} & \textbf{90.1} & \textbf{95.8} & \textbf{89.5} \\
\hline
\end{tabular}
\end{table*}

\begin{table*}[!t]
\caption{
Component-wise Ablation Study of the Proposed DSMCL on the CMShipReID Dataset Under Six Evaluation Protocols.
\textbf{Bold} values indicate the best performance.
\label{tab:ablation_ship}
}
\centering
\begin{tabular}{c|cc|cc|cc|cc|cc|cc|cc}
\hline 
\multirow{2}{*}{Method} & \multirow{2}{*}{SMCL} & \multirow{2}{*}{FDCL} & \multicolumn{2}{c|}{TIR to VIS}&\multicolumn{2}{c|}{VIS to TIR} & \multicolumn{2}{c|}{NIR to VIS} & \multicolumn{2}{c|}{VIS to NIR}& \multicolumn{2}{c|}{NIR to TIR}&\multicolumn{2}{c}{TIR to NIR}\\
\cline{4-15}
&&&Rank1&mAP&Rank1&mAP&Rank1&mAP&Rank1&mAP&Rank1&mAP&Rank1&mAP\\
\hline
TransReID & \ding{55} & \ding{55} & 95.6 & 66.8 & 91.4 & 73.4 & 76.9 & 57.0 & 72.7 & 57.2 & 75.1 & 59.2 & 73.2 & 54.4\\
TransReID + SMCL & \ding{51} & \ding{55} & 96.5 & 70.3 & 92.5 & 75.0 & 80.2 & 58.3 & 76.2 & 60.0 & 79.8 & 60.8 & 74.2 & 59.3 \\
TransReID + FDCL & \ding{55} & \ding{51} & 96.0 & 72.6 & 93.3 & 75.6 & 82.2 & 61.4 & 74.3 & 58.1 & 77.6 & 60.0 & 76.5 & 58.8 \\
TransReID + DSMCL & \ding{51} & \ding{51} & \textbf{96.6} & \textbf{73.9} & \textbf{94.0} & \textbf{75.4} & \textbf{83.8} & \textbf{63.2} & \textbf{79.8} & \textbf{60.9} & \textbf{78.1} & \textbf{61.4} & \textbf{77.4} & \textbf{60.2}\\
\hline
\end{tabular}
\end{table*}

\subsubsection{Results on Cross-modal Ship ReID Benchmarks}

Tables~\ref{tab:HOSS-ReID}, \ref{tab:CMShip1}, and \ref{tab:CMShip2} report the comparison results on HOSS-ReID and CMShipReID. Compared with visible-infrared person ReID, cross-modal ship retrieval is considerably more challenging due to the larger appearance discrepancy introduced by heterogeneous sensing modalities, including optical, SAR, near-infrared, and thermal infrared imagery.

On HOSS-ReID, DSMCL consistently improves all baseline models under the All-to-All, Optical-to-SAR, and SAR-to-Optical protocols. For example, IDKL+DSMCL improves the Rank-1 accuracy by 2.3\%, 3.1\%, and 7.5\% under the three protocols, respectively. When incorporated into MOS, DSMCL achieves the best overall performance, reaching 70.4\%, 41.5\%, and 47.8\% Rank-1 accuracy under the three evaluation settings. These results demonstrate the effectiveness of DSMCL in mitigating the substantial modality gap between optical and SAR imagery, further verifying its strong generalization capability across different sensor modalities.

On CMShipReID, DSMCL also achieves consistent improvements across all six cross-modal retrieval protocols. Significant gains are observed on TransReID, IDKL, MFENet, and HSFLNet, demonstrating that DSMCL is compatible with substantially different backbone architectures. Moreover, consistent performance gains are observed across all baseline models, and several DSMCL-enhanced variants achieve state-of-the-art results under multiple evaluation protocols. This further validates the effectiveness of the proposed framework in complex multi-modal retrieval scenarios involving visible, near-infrared, and thermal infrared modalities.

Overall, DSMCL consistently improves retrieval performance across five benchmark datasets, seventeen evaluation protocols, and multiple representative baseline architectures. These results demonstrate the universal applicability of the proposed framework across diverse cross-modal scenarios and verify its ability to automatically capture modality-specific characteristics and modality-shared identity cues for robust representation learning.

\subsection{Ablation Study}

\subsubsection{Component Analysis on Visible-Infrared Person ReID}

Table~\ref{tab:ablation_sysu_and_regcb} reports the component-wise ablation results on SYSU-MM01 and RegDB. Compared with the baseline IDKL, introducing SMCL alone consistently improves retrieval performance across all evaluation protocols, demonstrating the effectiveness of spatial modality consistency learning in reducing cross-modal feature distribution discrepancy. Similarly, incorporating FDCL alone also yields notable performance gains, indicating that explicitly modeling high-frequency discriminative representations provides complementary modality-invariant information beyond conventional spatial-domain learning for cross-modal retrieval tasks.

When both SMCL and FDCL are jointly incorporated into the baseline model, the resulting DSMCL-enhanced variant consistently achieves the best performance under all evaluation protocols. For example, on SYSU-MM01 All-search Single-shot, the Rank-1 accuracy and mAP are improved from 80.9\% and 78.5\% to 83.6\% and 81.4\%, respectively. Similar improvements are observed on RegDB, where IDKL enhanced by DSMCL achieves 95.4\%/90.1\% and 95.8\%/89.5\% Rank-1/mAP under the Visible-to-Infrared and Infrared-to-Visible settings, respectively. These results demonstrate that SMCL and FDCL provide complementary benefits, and jointly incorporating both modules consistently yields the most discriminative and modality-robust representations.

\subsubsection{Component Analysis on Cross-modal Ship ReID}

Table~\ref{tab:ablation_ship} presents the ablation results on the CMShipReID benchmark in our study. Similar observations can be made across all six evaluation protocols. Both SMCL and FDCL individually improve the baseline performance, confirming the effectiveness of spatial consistency learning and frequency-aware discriminative consistency learning in multi-modal maritime retrieval scenarios.

Furthermore, jointly incorporating SMCL and FDCL consistently yields the best performance under all evaluation protocols. For instance, introducing both modules improves the Rank-1 accuracy of TransReID from 91.4\% to 94.0\% under the VIS-to-TIR protocol and from 72.7\% to 79.8\% under the VIS-to-NIR protocol. Compared with introducing only SMCL or FDCL, incorporating both modules simultaneously consistently achieves superior Rank-1 accuracy and mAP across all test settings. These results further confirm that spatial-domain modality alignment and frequency-domain discriminative consistency are complementary, and their joint optimization consistently leads to the best retrieval performance across diverse maritime sensing scenarios.

\subsection{Hyperparameter Sensitivity Analysis}

\begin{figure}[!t]
\centering
\includegraphics[width=3.5in]{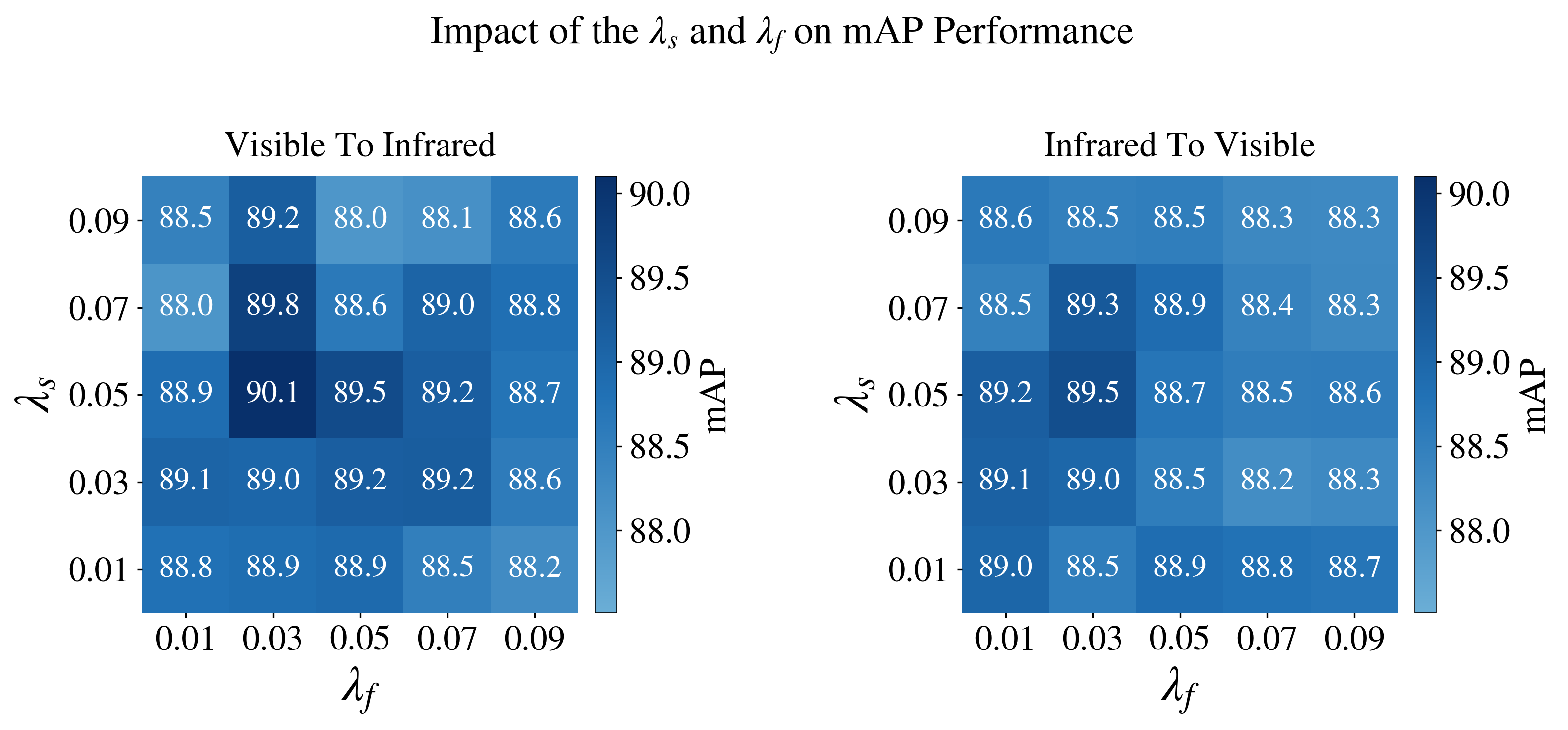}
\caption{Impact of the spatial consistency weight $\lambda_s$ and frequency consistency weight $\lambda_f$ on mAP under the Visible-to-Infrared and Infrared-to-Visible settings of RegDB.}
\label{fig:heatmap}
\end{figure}

\begin{figure}[!t]
\centering
\includegraphics[width=3.5in]{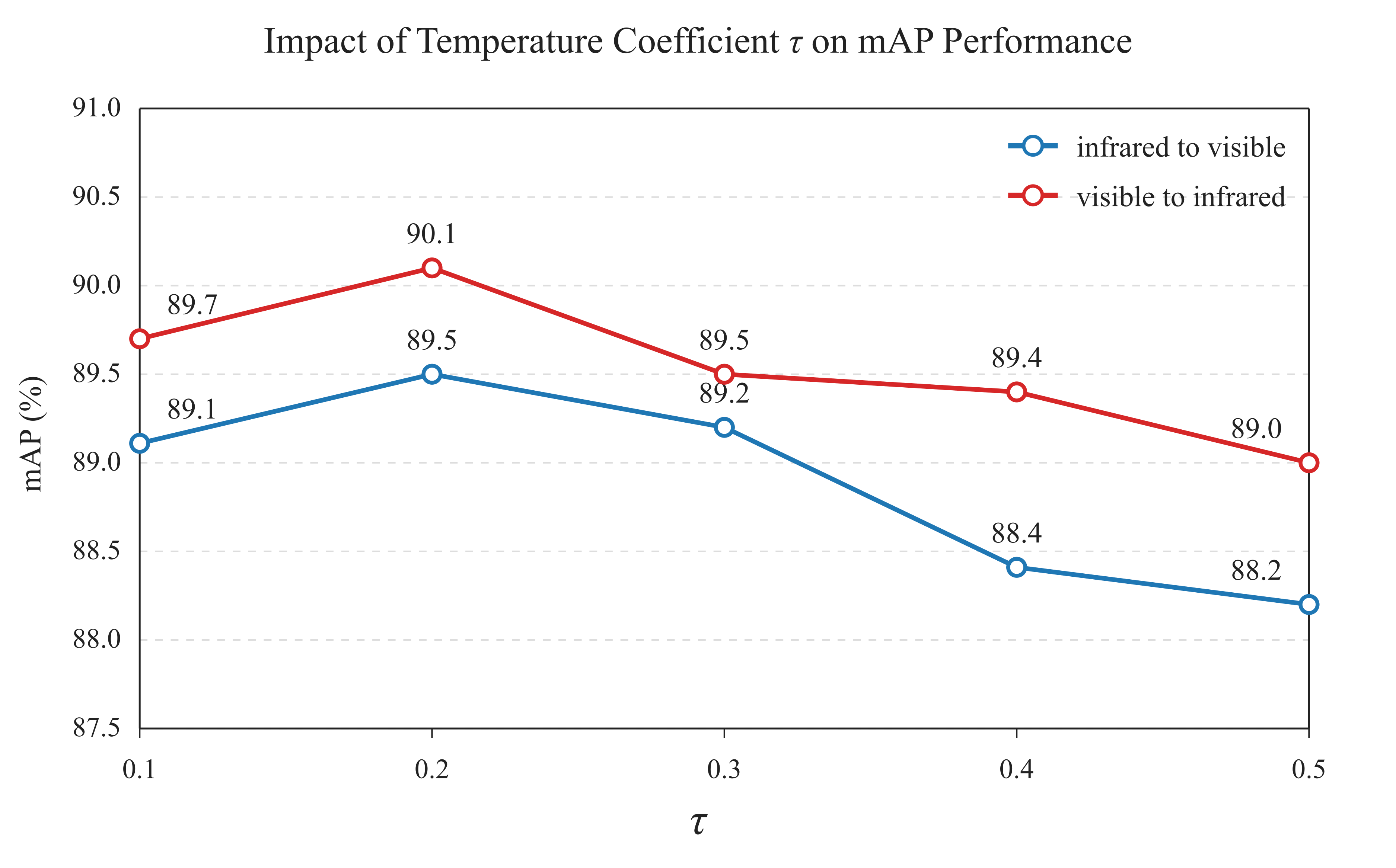}
\caption{Sensitivity analysis of the temperature parameter $\tau$ on RegDB under the Visible-to-Infrared and Infrared-to-Visible settings of RegDB.}
\label{fig:tau}
\end{figure}

To investigate the influence of the proposed loss weights and contrastive temperature parameter, we conduct hyperparameter sensitivity experiments on the RegDB dataset under both Visible-to-Infrared and Infrared-to-Visible evaluation protocols. We select IDKL as baseline method.

Fig.~\ref{fig:heatmap} illustrates the impact of the spatial consistency weight $\lambda_s$ and frequency consistency weight $\lambda_f$. We vary both parameters from 0.01 to 0.09 while keeping $\tau = 0.2$. It can be observed that the performance remains relatively stable across a broad range of parameter combinations, indicating the robustness of the proposed DSMCL framework. The best performance is consistently achieved when $\lambda_s=0.05$ and $\lambda_f=0.03$, yielding 90.1\% and 89.5\% mAP under the Visible-to-Infrared and Infrared-to-Visible settings, respectively. These results suggest that an appropriate balance between spatial modality alignment and frequency-aware discriminative learning is beneficial for learning robust representations.

Fig.~\ref{fig:tau} presents the sensitivity analysis of the temperature parameter $\tau$ used in the frequency-aware discriminative contrastive learning objective. As $\tau$ increases from 0.1 to 0.2, the performance gradually improves and reaches its optimum at $\tau=0.2$. Further increasing $\tau$ leads to a noticeable performance decline under both evaluation protocols. This phenomenon can be attributed to the trade-off between sample discrimination and representation smoothness in the learned feature space. A small temperature may overly emphasize hard samples, while a large temperature weakens the discriminative capability of the contrastive objective. Therefore, $\tau=0.2$ is adopted in all experiments.

Overall, the results demonstrate that DSMCL is not overly sensitive to hyperparameter selection and achieves consistently strong performance within a relatively wide parameter range.

\subsection{Visualization Analysis}

\subsubsection{t-SNE and Distance Distribution Analysis}

To further investigate the effectiveness of the proposed DSMCL, we conduct both qualitative and quantitative analyses on the learned feature representations. Specifically, ten identities are randomly selected from the RegDB test set for visualization. We select HSFLNet as baseline method.

\begin{figure}[!t]
\centering
\includegraphics[width=3.5in]{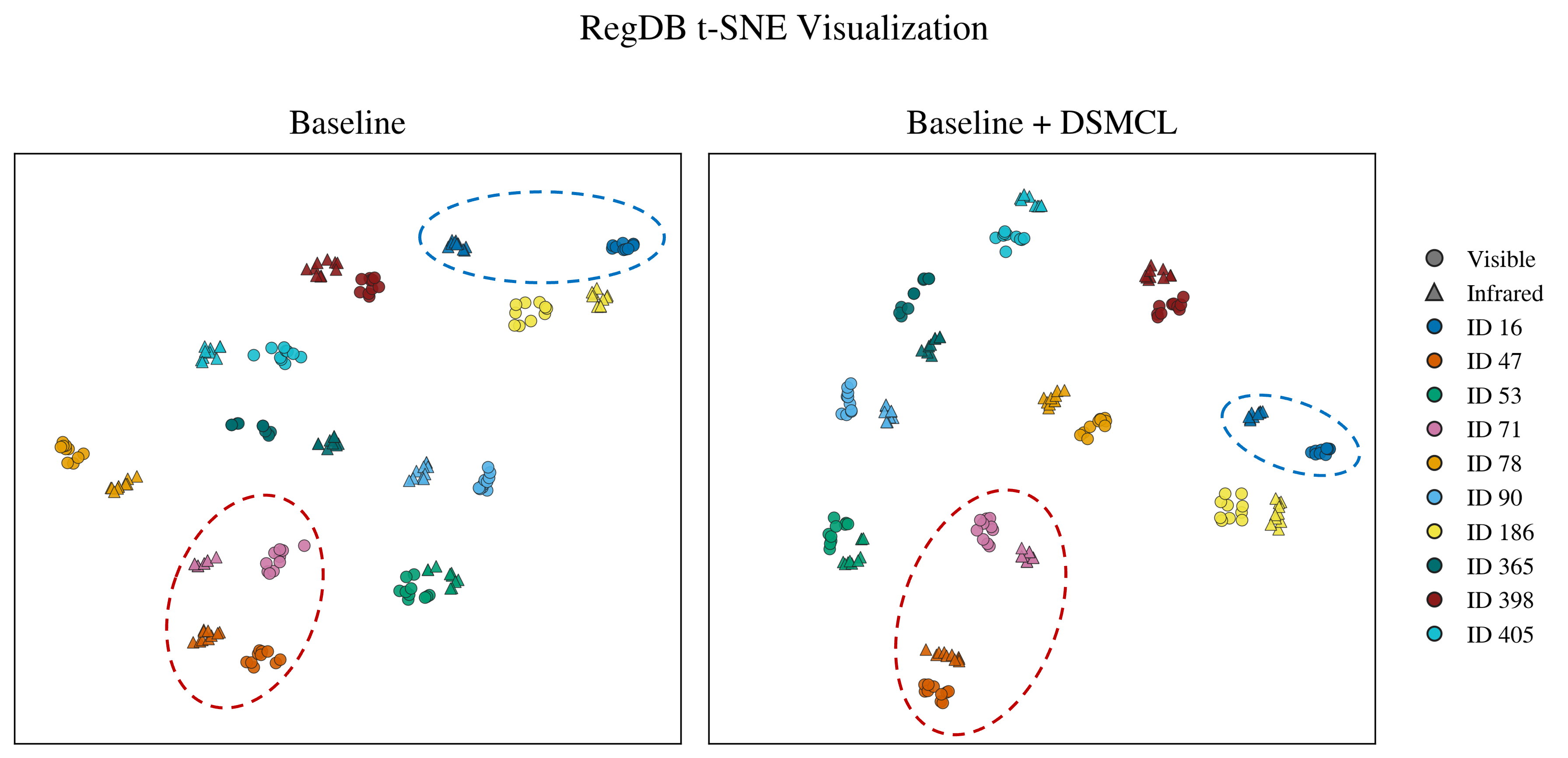}
\caption{T-SNE visualization of feature embeddings for randomly selected identities from the RegDB dataset. Different colors denote different identities, while circles and triangles represent visible and infrared samples, respectively.}
\label{fig:tsne}
\end{figure}

Fig.~\ref{fig:tsne} presents the t-SNE visualization of the learned embeddings. Compared with the baseline model, DSMCL produces more compact intra-class distributions and clearer separation between different identities. As highlighted by the blue dashed ellipse, samples belonging to the same identity are clustered more closely after introducing DSMCL. Meanwhile, the red dashed ellipse demonstrates that DSMCL effectively alleviates the confusion between neighboring identity clusters, resulting in improved feature discrimination.

\begin{figure}[!t]
\centering
\includegraphics[width=3.5in]{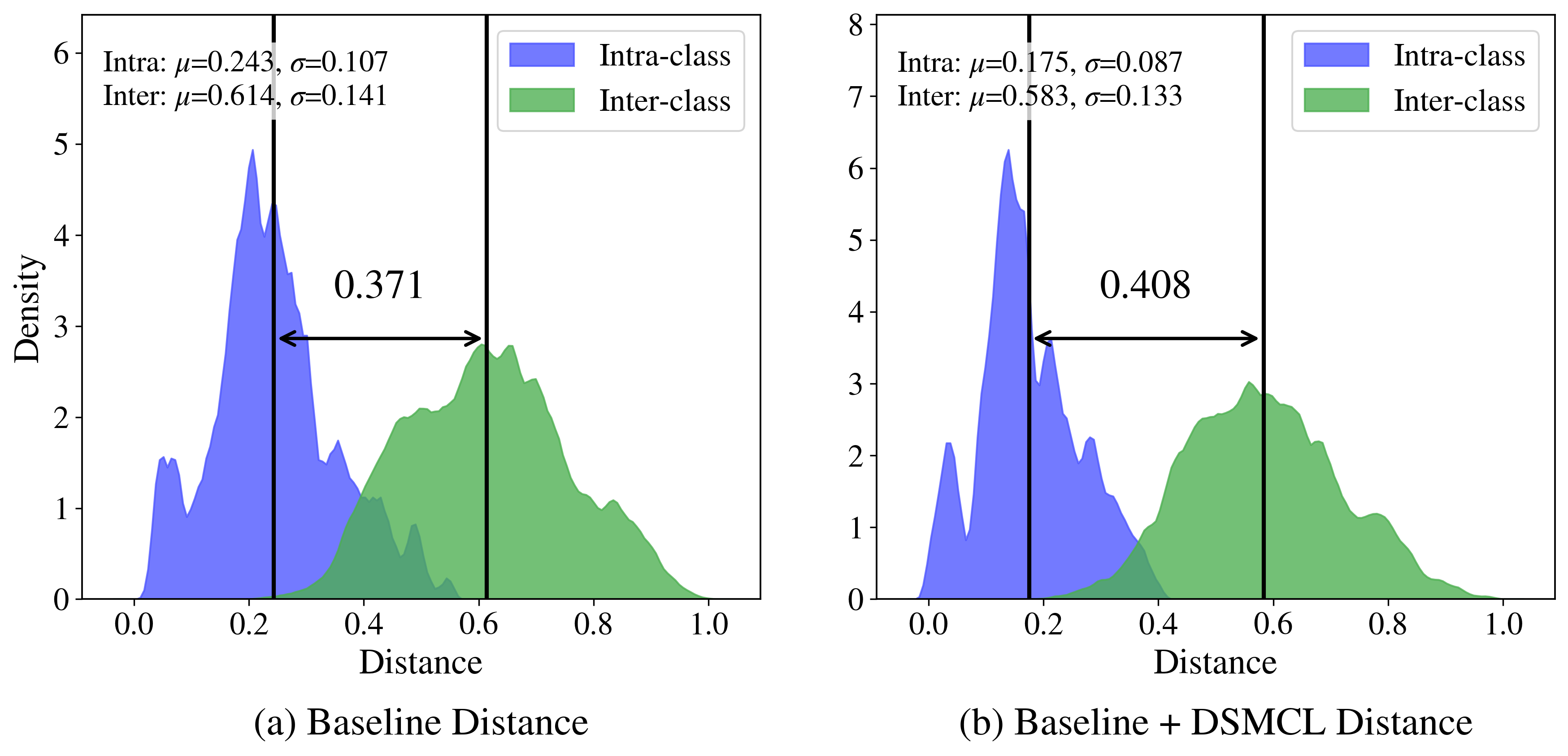}
\caption{Probability density distributions of intra-class and inter-class feature distances computed from the same randomly selected RegDB samples used in Fig.~\ref{fig:tsne}. The blue and green regions denote the intra-class and inter-class distance distributions, respectively.}
\label{fig:distance}
\end{figure}

To quantitatively validate these observations, Fig.~\ref{fig:distance} further illustrates the probability density distributions of intra-class and inter-class distances for the same identities. Compared with the baseline, DSMCL reduces the average intra-class distance from 0.243 to 0.175 while maintaining a relatively large inter-class distance. Consequently, the distance margin between inter-class and intra-class distributions increases from 0.371 to 0.408, demonstrating a clear improvement in feature separability. This enlarged margin indicates that DSMCL learns more compact representations for samples of the same identity while preserving stronger discrimination between different identities.

Overall, the visualization results consistently demonstrate that the proposed dual-space modality consistency learning effectively enhances intra-class compactness and inter-class separability, leading to more discriminative and accurate feature representations across diverse modalities.

\subsubsection{Frequency Spectrum Alignment Analysis}

\begin{figure}[!t]
\centering
\includegraphics[width=3.5in]{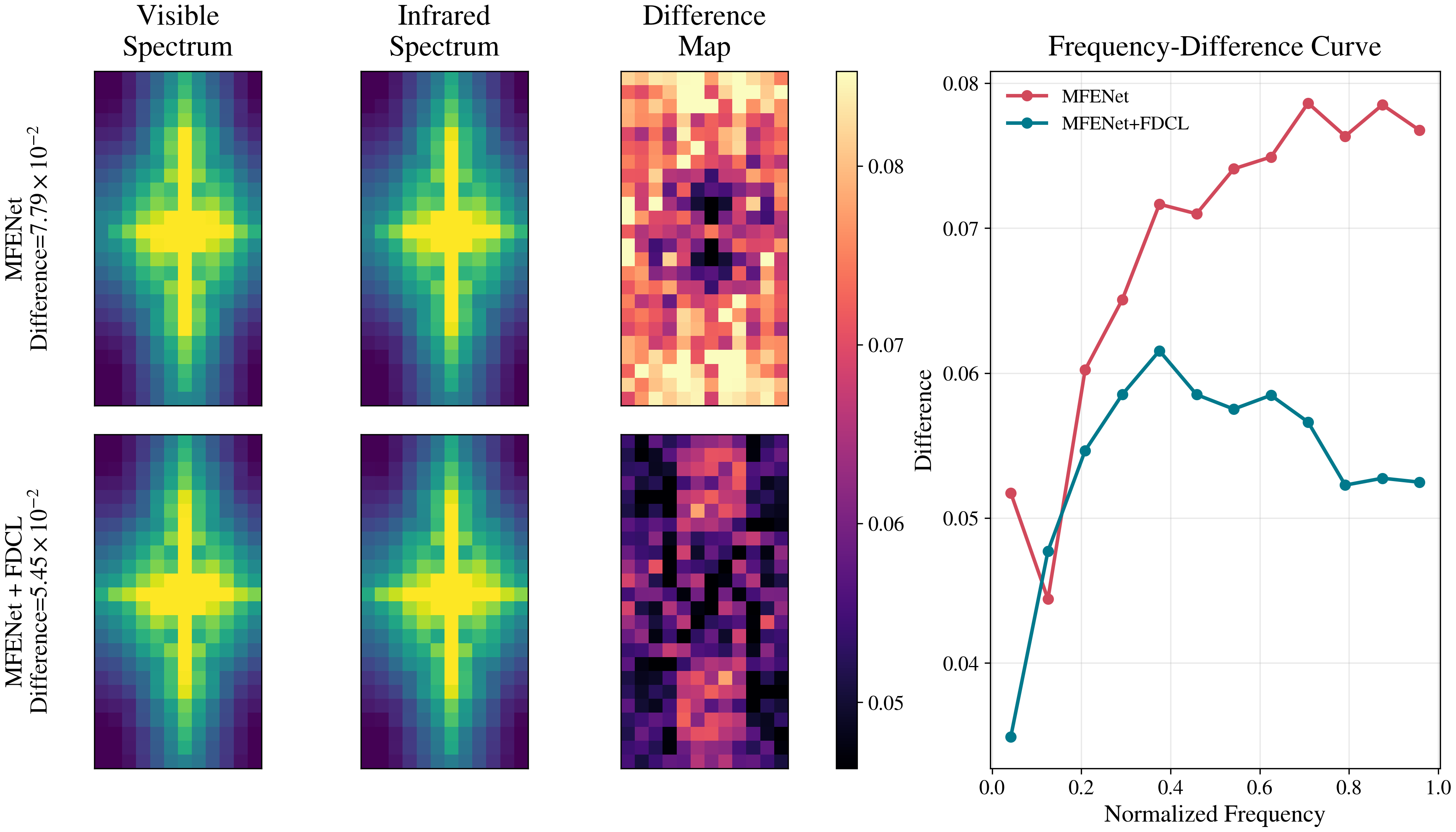}
\caption{Cross-modal frequency spectrum alignment analysis for randomly selected identities from the SYSU-MM01 dataset. Lower spectrum discrepancy scores indicate better frequency-domain consistency across modalities.}
\label{fig:sysu_spectrum_alignment}
\end{figure}

To further verify the effectiveness of FDCL in the frequency domain, we select MFENet as the baseline method and analyze cross-modal frequency spectrum alignment on the SYSU-MM01 and CMShipReID datasets. Specifically, ten identities are randomly selected from the test set. For each identity, intermediate feature maps are transformed into the frequency domain via FFT, and the averaged spectra of different modalities are computed. The absolute difference between paired spectra is then used to quantify cross-modal frequency discrepancy.

\begin{figure}[!t]
\centering
\includegraphics[width=3.5in]{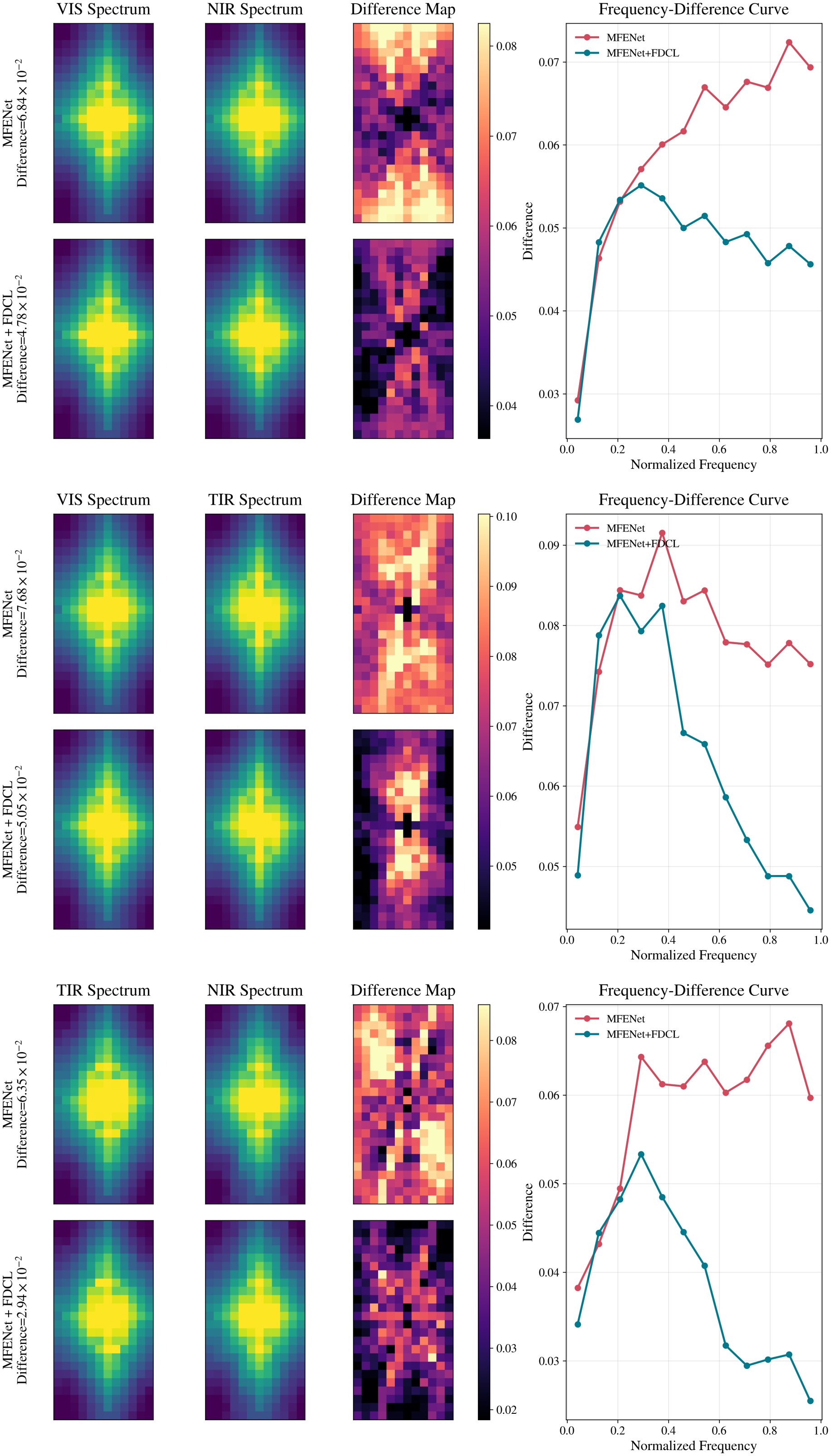}
\caption{Cross-modal frequency spectrum alignment analysis for randomly selected identities from the CMShipReID dataset. Lower spectrum discrepancy scores indicate better frequency-domain consistency across modalities.}
\label{fig:cmship_spectrum_alignment}
\end{figure}
As shown in Fig.~\ref{fig:sysu_spectrum_alignment}, the frequency spectra are visualized after applying $\mathrm{FFTshift}(\cdot)$, where low-frequency components are concentrated around the center and high-frequency components are distributed toward the periphery in the frequency domain. Since the energy of most visual representations is predominantly concentrated in low-frequency regions, the overall spectral structures of visible and infrared modalities appear visually similar for both MFENet and MFENet+FDCL. Therefore, the global spectrum distributions alone are insufficient to reveal subtle modality discrepancies.

To better characterize frequency-domain consistency, we further analyze the difference maps, spectrum discrepancy scores, and frequency-difference curves. Here, the spectrum discrepancy score is defined as the mean value of the corresponding difference map, where a lower value indicates better frequency-domain consistency across modalities . Compared with the MFENet baseline, MFENet+FDCL produces noticeably lower responses in the difference map, indicating reduced spectral discrepancy between visible and infrared modalities. Quantitatively, the discrepancy score decreases from 7.79$\times 10^{-2}$ to 5.45$\times 10^{-2}$, corresponding to a relative reduction of approximately 30.0\%.

Moreover, the frequency-difference curve reveals that the improvement is primarily concentrated in high-frequency regions, whereas the discrepancy in low-frequency components remains relatively stable. This observation is consistent with our motivation that high-frequency representations are both highly discriminative and highly modality-sensitive. While low-frequency components mainly preserve shared semantic structures across modalities, high-frequency components contain richer modality-specific variations and therefore contribute more substantially to cross-modal discrepancy. By explicitly enforcing high-frequency consistency across modalities, FDCL effectively suppresses modality-specific frequency variations while preserving identity-related discriminative information. 

\begin{figure}[!t]
\centering
\includegraphics[width=3.5in]{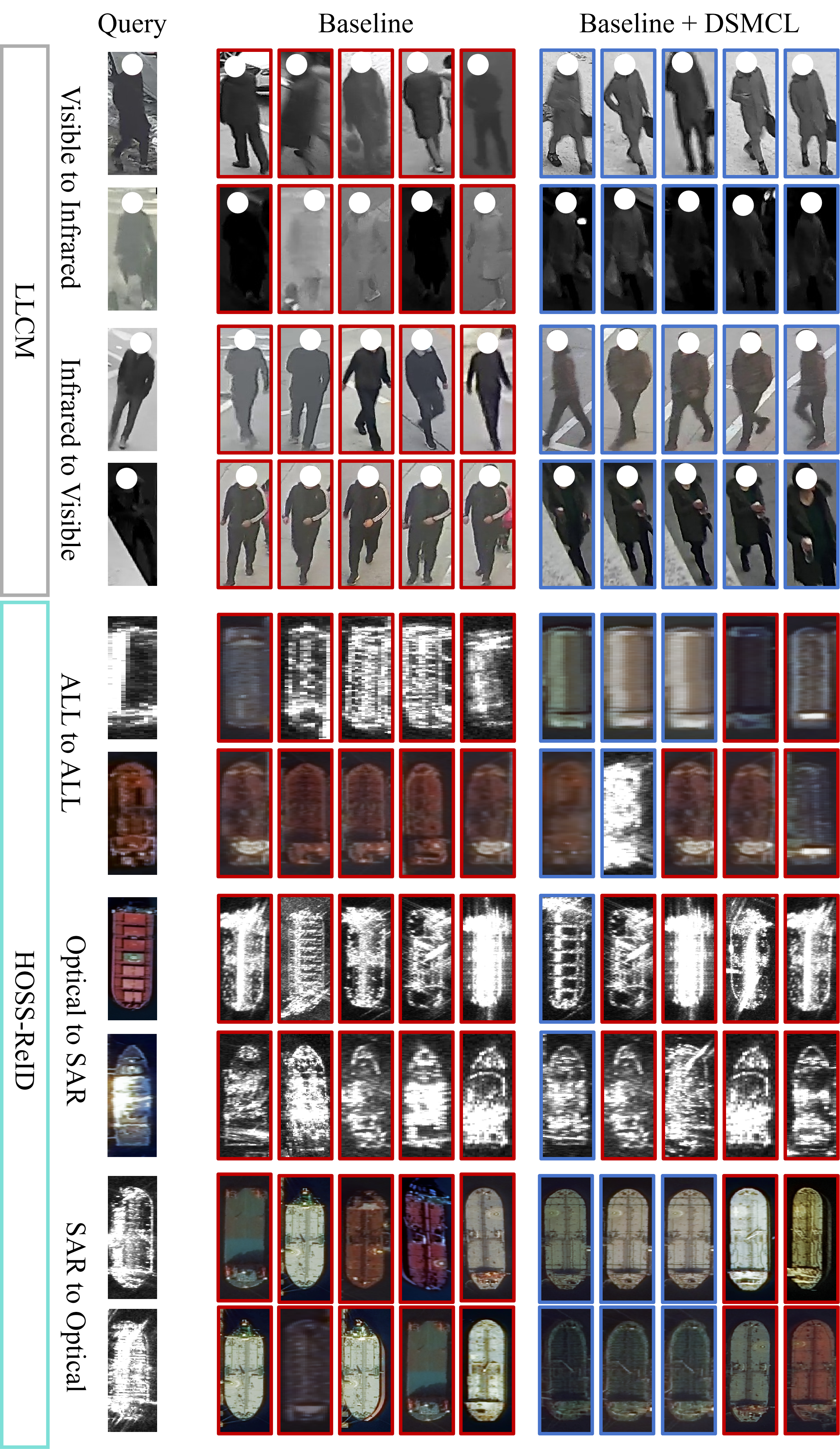}
\caption{Qualitative retrieval comparison between between the IDKL baseline and the baseline augmented with DSMCL for randomly selected queries from the LLCM and HOSS-ReID datasets. For each query image, the top-5 retrieved gallery samples are presented. Blue and red boxes indicate correct and incorrect matches, respectively.}
\label{fig:top5}
\end{figure}


To evaluate the effectiveness of FDCL under multi-modal settings, we conduct frequency spectrum alignment analysis on the CMShipReID dataset. As shown in Fig.~\ref{fig:cmship_spectrum_alignment}, FDCL reduces spectral discrepancy across all modality pairs, including VIS-NIR, VIS-TIR, and NIR-TIR. Compared with the MFENet baseline, the difference maps exhibit lower responses with FDCL. The spectrum discrepancy scores are reduced from 6.84$\times 10^{-2}$ to 4.78$\times 10^{-2}$ for VIS-NIR, from 7.68$\times 10^{-2}$ to 5.05$\times 10^{-2}$ for VIS-TIR, and from 6.35$\times 10^{-2}$ to 2.94$\times 10^{-2}$ for NIR-TIR, respectively.


More importantly, the frequency-difference curves demonstrate that the improvement is mainly concentrated in high-frequency regions. While low-frequency components preserve common semantic structures shared across modalities, high-frequency components contain richer modality-specific characteristics and therefore contribute more substantially to cross-modal discrepancy. The consistent reduction of high-frequency differences across all modality pairs indicates that FDCL effectively aligns modality-specific frequency representations and promotes frequency-domain consistency. This behavior suggests that FDCL selectively regularizes modality-sensitive spectral components without unnecessarily disturbing shared low-frequency structures. These results verify that FDCL consistently reduces cross-modal frequency discrepancy across both visible-infrared and multi-modal ReID scenarios.


Notably, the observed improvements are achieved on top of the MFENet baseline, which already incorporates frequency-domain modeling for feature representation. This suggests that the performance gain of FDCL does not merely stem from exploiting frequency information itself, but rather from explicitly enforcing cross-modal frequency consistency. In other words, frequency representation learning and cross-modal frequency consistency address distinct yet complementary objectives. These observations provide direct evidence for the effectiveness of the proposed frequency-aware discriminative consistency learning strategy and demonstrate its complementary benefits to existing frequency-aware ReID frameworks.

\subsubsection{Top-5 Retrieval Analysis} To provide a more intuitive evaluation of retrieval performance, Fig.~\ref{fig:top5} presents qualitative retrieval comparisons on the LLCM and HOSS-ReID datasets, where the IDKL baseline is compared with the baseline augmented by the proposed DSMCL.


For the LLCM dataset, incorporating DSMCL consistently improves the retrieval quality of the baseline. In particular, all top-5 retrieved samples correspond to the same identity for the selected queries, demonstrating enhanced cross-modal discriminability compared with the original IDKL model. Moreover, LLCM contains significant low-light degradation and severe appearance ambiguity, which substantially increase the difficulty of cross-modal retrieval. Despite these challenges, the baseline augmented with DSMCL still maintains highly accurate retrieval results. Compared with the baseline, DSMCL also reduces false matches involving visually similar identities. These observations indicate that dual-space modality consistency learning effectively separates identity-discriminative cues from modality-induced variations under adverse illumination conditions. 

For the HOSS-ReID dataset, the introduction of DSMCL also brings clear improvements over the baseline. All correct gallery samples belonging to the queried identity are successfully retrieved after incorporating DSMCL. Notably, the first query corresponds to a low-resolution ship sample under the ALL-to-ALL evaluation protocol, where the enhanced model still achieves accurate retrieval despite the limited visual details. Furthermore, during modality-mixed retrieval, the baseline augmented with DSMCL is able to simultaneously retrieve gallery samples from both modalities that belong to the same identity. This observation suggests that DSMCL effectively reduces modality discrepancy while preserving identity consistency, thereby bringing heterogeneous modality representations closer in the shared feature space.

Overall, the qualitative retrieval results are consistent with the quantitative improvements reported in the benchmark evaluations. In particular, DSMCL reduces mismatches caused by severe illumination variations, limited visual details, and heterogeneous sensing characteristics while preserving identity-relevant cues. Its consistent benefits across person and ship retrieval further support its applicability to distinct target categories and modality combinations. These results demonstrate that DSMCL serves as an effective plug-and-play enhancement for existing cross-modal ReID frameworks, improving cross-modal discriminability, retrieval robustness, and alignment between heterogeneous modalities. 

\section{Conclusion}

In this paper, we proposed a novel Dual-Space Modality Consistency Learning framework (DSMCL) for universal cross-modal re-identification. Unlike existing methods that are typically designed for specific modality combinations, DSMCL serves as a unified and plug-and-play modality consistency learning framework that can be readily integrated into different baseline architectures and cross-modal scenarios. Specifically, DSMCL jointly exploits spatial-domain modality consistency and frequency-domain discriminative consistency through the proposed SMCL and FDCL modules, thereby reducing modality discrepancy while preserving discriminative identity information.

Extensive experiments on five benchmark datasets, including SYSU-MM01, RegDB, LLCM, HOSS-ReID, and CMShipReID, demonstrate the effectiveness of the proposed framework. DSMCL consistently improves multiple representative baseline models across seventeen evaluation protocols spanning visible-infrared person ReID, optical-SAR ship ReID, and multi-modal ship ReID scenarios. These results validate the strong versatility and generalization capability of DSMCL across substantially different sensing modalities without requiring modality-specific architectural designs.

In future work, we will explore more adaptive modality-aware consistency learning strategies and extend the proposed framework to broader cross-modal retrieval tasks.




\bibliographystyle{IEEEtran}
\bibliography{IEEEabrv,reference}

\vspace{11pt}

\vfill

\end{document}